%% file: paper.tex
\documentclass{article}

\usepackage[preprint]{neurips_2026}
\usepackage[utf8]{inputenc}
\usepackage[T1]{fontenc}
\usepackage{hyperref}
\usepackage{url}
\usepackage{booktabs}
\usepackage{amsfonts}
\usepackage{amsmath}
\usepackage{nicefrac}
\usepackage{microtype}
\usepackage{xcolor}
\usepackage{graphicx}
\usepackage{subcaption}
\usepackage{multirow}
\usepackage{enumitem}
\usepackage{wrapfig}
\usepackage{float}

\newcommand{\best}[1]{\textbf{#1}}
\newcommand{\srdrn}{\textsc{srdrn}}
\newcommand{\qsrdrn}{Q-\textsc{srdrn}}
\newcommand{\cvae}{c\textsc{vae}}

\title{Multi-Quantile Regression for Extreme \\ Precipitation Downscaling}

\author{
  Hamed Najafi\\
  Florida International University \\
  Miami, FL \\
  \texttt{hnaja002@fiu.edu} \\
  \AND
  Gareth Lagerwall \\
  Everglade foundation \\
  Miami, FL \\
  \texttt{glagerwall@evergladesfoundation.org} \\ 
  \AND
  Jayantha Obeysekera \\
  Florida International University \\
  Miami, FL \\
  \texttt{jobeysek@fiu.edu} \\ 
  \AND
  Jason Liu\\
  Florida International University \\
  Miami, FL \\
  \texttt{liux@fiu.edu} \\
}
\begin{document}

\maketitle

\begin{abstract}

Deep super-resolution networks for precipitation downscaling achieve strong bulk
skill yet systematically under-predict the heavy-tail events that drive flood
risk. The natural remedy---augmenting training with synthetic extremes---turns
out fruitless. We argue that the obstacle is the loss, not the data: under
intensity-weighted MAE, real and synthetic labels at the same input are simply
averaged, so augmentation shifts the predicted mean rather than the conditional
distribution. We resolve this with \qsrdrn{}, a multi-quantile super-resolution
network trained with pinball loss at $\tau \in \{0.50, 0.95, 0.99, 0.999\}$. Two
CNN-specific design choices make this practical: \emph{IncrementBound} enforces
monotonicity while preserving each quantile channel's gradient identity, and
separate per-quantile output heads give the bulk and tail detectors independent
filter banks. Under this design, data augmentation (\cvae{}) becomes complementary
rather than harmful: the median head absorbs synthetic extreme patterns without
contaminating the upper quantiles. Empirically, on Florida (convective, tropical-cyclone-dominated) the
un-augmented \qsrdrn{} P999 head detects $1{,}598$ of $2{,}111$ events at
$200$\,mm/day versus $88$ for the deterministic baseline---an $18{\times}$
detection-rate gain ($4.2\% \to 75.7\%$)---with $63\%$ lower KL divergence and
$3.9\%$ lower RMSE. Additionally, $83$ \cvae{}-generated samples lift the P50 channel
from $14$ to $1{,}038$ hits at $200$\,mm/day ($74{\times}$). On California (atmospheric-river-dominated), the
architecture alone reaches near-perfect detection (P999 SEDI $\ge 0.996$
through $300$\,mm/day).  On a Texas substate, the architecture-only effect
sharpens further: the deterministic baseline only catches $2$ of $10{,}720$ events
at $200$\,mm/day while the P999 head catches $8{,}776$ ($81.9\%$). Interestingly, the FL-tuned
\cvae{} does not transfer to California or the Texas substate, exposing
augmentation itself as a region-specific component. 
Multi-quantile regression captures extremes wherever the
large-scale signal is strong; \cvae{} augmentation, when its architecture 
matches to the regime, rescues the median where it is not.

\end{abstract}

\section{Introduction}
\label{sec:intro}

Precipitation downscaling, which maps coarse reanalysis fields to high-resolution
observations, is a routine input to flood-risk modeling, infrastructure planning,
and operational hydrology. Modern super-resolution networks match the bulk of
observed precipitation distributions well~\citep{vandal2017deepsd, sha2020deep,
wang2021deep}, yet their predictions degrade systematically at the heaviest events
that matter most for downstream risk. On Florida ERA5/PRISM data, a
state-of-the-art deterministic baseline~\citep{wang2021deep} over-predicts the
climatological mean by $43\%$ ($5.94$ vs.\ $4.14$\,mm/day) while detecting only
$88$ of $2{,}111$ observed events at $\ge200$\,mm/day ($4.2\%$ probability of detection, POD). The
textbook fix---enlarging the training set with synthetic extremes from a generative
model---also fails: performance degrades rather than improves as augmentation grows.

The failure is a property of the loss, not the data.  An intensity-weighted
MAE converges to a weighted conditional median~\citep{koenker2001quantile}:
at a pixel where $99\%$ of days are light, the optimal prediction sits near
the bulk, and a synthetic extreme is simply averaged into the median.  The
natural alternative is pinball-loss multi-quantile
regression~\citep{koenker2001quantile, bremnes2004probabilistic}, which
assigns one output per quantile level and penalizes under-prediction
$\tau/(1-\tau)$ times more heavily than over-prediction (a $999{\times}$
asymmetry at $\tau{=}0.999$).  

Two CNN-specific obstacles must be addressed
simultaneously: per-pixel sorting (the textbook non-crossing fix) scrambles
the gradient identity of each output channel so no convolutional filter
specializes; and a shared \texttt{Conv2D(4)} output layer forces bulk and
tail detectors to compete for the same filter bank.  We resolve both with
\emph{IncrementBound}, a cumulative-softplus construction that pins each
output channel to a fixed quantile, combined with \emph{separate}
\texttt{Conv2D(1)} heads.  An unexpected payoff is that the P50 head, trained
with unweighted pinball loss, converges to the true conditional median and
is robust to a small fraction of synthetic extremes---so multi-quantile regression
and augmentation, orthogonal in isolation, become complementary.  

We
validate on three U.S.\ domains: Florida (convective, tropical-cyclone),
California (atmospheric-river), and a Texas Gulf-Coast substate (mixed
convective/tropical-cyclone landfall).  \cvae{} augmentation is only reported on
Florida because the same FL-tuned generator architecture mildly harms
California's P50 channel and was therefore not run on the Texas substate---a
finding we treat as a separate result on the regime-specificity of
augmentation (\S\ref{subsec:augmentation}).

Overall, our contributions are three-folds:
{\bf\em First,} we identify a CNN-specific failure mode in sort-based quantile
        monotonicity---per-pixel permutations destroy spatial filter
        specialization---and resolve it with \qsrdrn{}, which combines
        IncrementBound with separate \texttt{Conv2D(1)} heads. Without any
        augmentation, P999 detection at $200$\,mm/day on Florida rises from
        $4.2\%$ to $75.7\%$ POD ($88$ to $1{,}598$ events).
 {\bf\em Second,} we validate the same architecture and hyperparameters across three
        climatologically distinct domains.  On California
        (atmospheric-river-dominated) the un-augmented architecture reaches
        P999 SEDI $\ge 0.996$ at every threshold up to $300$\,mm/day; on a
        Texas substate the un-augmented P999 head catches $8{,}776$ of
        $10{,}720$ events at $200$\,mm/day ($81.9\%$, vs.\ $0.02\%$ for the
        deterministic baseline) and $1{,}441$ of $2{,}265$ at $300$\,mm/day.
 {\bf\em Finally,} we show that multi-quantile regression is what makes \cvae{} augmentation
        useful on Florida: under pinball loss the median is shielded from the
        label-averaging that poisons MAE-trained networks, lifting Florida P50
        SEDI at $200$\,mm/day from $0.405$ to $0.866$ and reducing KL divergence
        by $50\%$.  

\section{Related Work}
\label{sec:related}

{\bf Statistical and learning-based downscaling.}
Statistical downscalers---bias correction and spatial
disaggregation~\citep{wood2004hydrologic}, generalized linear
models~\citep{chandler2007bias}, and analog
methods~\citep{zorita1999analog}---are calibrated by construction but cannot
exploit high-dimensional reanalysis fields. Deep learning approaches, including
CNNs~\citep{vandal2017deepsd, bano2020configuration}, super-resolution residual
networks~\citep{wang2021deep}, and U-Nets~\citep{sha2020deep}, lift bulk skill
substantially but inherit the mean-seeking bias of their training
objectives---whether pixel-wise MAE/MSE point losses or expected-value
parametric likelihoods (e.g., the Bernoulli--gamma NLL used for precipitation
in \citealp{bano2020configuration})---the very failure mode this paper
diagnoses. Stochastic extensions
sidestep mean-seeking by sampling rather than by changing the loss:
GANs~\citep{harris2022generative, price2022increasing} produce ensemble samples,
and diffusion-based samplers---whether for km-scale regional downscaling
(CorrDiff~\citep{mardani2025residual}, $2$\,km Taiwan) or global ensemble
forecasting (GenCast~\citep{price2025gencast}, $0.25^{\circ}$
latitude--longitude)---extend ensemble sampling to calibrated probabilistic
distributions at the cost of an iterative multi-step generation process per
realization (roughly $12$ denoising iterations for CorrDiff and $39$ network
evaluations per timestep for GenCast). Foundation-scale weather
models~\citep{pathak2022fourcastnet, bi2023pangu, nguyen2023climax} target
global forecasting and are complementary in scope. \qsrdrn{} differs from the
sampling-based family by replacing the loss: a single forward pass returns a
calibrated discrete CDF at every pixel, including a dedicated extreme-tail
quantile.

{\bf Quantile regression and monotonic networks.}
Pinball-loss quantile regression for precipitation has a long
history~\citep{bremnes2004probabilistic, cannon2011quantile}. Non-crossing has
been enforced through several mechanisms: cumulative softplus increments
between successive quantile heads~\citep{cannon2018non}, structurally monotonic
feed-forward networks built from positive weights or integrated positive
functions~\citep{daniels2010monotone, wehenkel2019unconstrained,
brando2022deep}, and constrained quantile increments inside recurrent
forecasters~\citep{liu2023novel}. Each of
these approaches assumes a one-to-one mapping between output unit and quantile,
which holds in fully-connected and recurrent decoders but breaks in spatial
CNNs, where a single convolutional filter is reused at every pixel. We diagnose
the resulting CNN-specific failure mode---per-pixel sort permutations scramble
the gradient identity of each channel (Sec.~\ref{subsec:ablation})---and
resolve it with IncrementBound combined with separate \texttt{Conv2D(1)} output
heads, which together preserve a fixed gradient-to-channel routing.
Sampling-based uncertainty estimators~\citep{gal2016dropout,
lakshminarayanan2017simple, vandal2018uncertainty} are a probabilistic
alternative but, as our comparisons show (\S\ref{subsec:uq_baselines}), collapse
to a quantile-invariant predictive variance at the precipitation tails.

{\bf Data augmentation for imbalanced geophysical data.}
Variational autoencoders~\citep{kingma2014auto} and their conditional
extensions~\citep{sohn2015learning}, alongside GANs, have been used to
synthesize weather fields~\citep{ravuri2021skilful},
but augmenting training sets with such samples to address the under-represented
precipitation tail remains underexplored. Existing work treats the generator as
a drop-in data source independent of the supervised loss. We show that the
\emph{interaction} between augmented samples and the loss---rather than the
quality of the synthetic samples---determines whether augmentation helps or
hurts: the same \cvae{} samples that destabilize an MAE-trained network are
absorbed by a calibrated multi-quantile predictor without contaminating the
upper quantile heads. Architecture and augmentation are therefore inseparable:
the former determines whether the latter helps or harms.

\section{Method}
\label{sec:method}

\subsection{Problem Setup}
\label{subsec:setup}

\begin{wraptable}{r}{0.66\textwidth}
\vspace{-0.9\baselineskip}
\centering
\caption{All three study domains share the same 15 ERA5 input variables over
${\sim}$12{,}400 training days (1980--2013).  Texas substate covers a
$6^\circ\times 6^\circ$ Gulf-Coast box ($28$--$34^\circ$N, $98$--$92^\circ$W).}
\label{tab:domains}
\vspace{2pt}
\footnotesize
\setlength{\tabcolsep}{4pt}
\begin{tabular}{lccc}
\toprule
\textbf{Attribute} & \textbf{Florida} & \textbf{California} & \textbf{Texas substate} \\
\midrule
ERA5 grid ($H \times W$) & $26 \times 33$ & $44 \times 48$ & $24 \times 24$ \\
PRISM grid ($H' \times W'$) & $156 \times 132$ & $264 \times 288$ & $144 \times 144$ \\
Output pixels & 20{,}592 & 76{,}032 & 20{,}736 \\
Land fraction & ${\sim}$40\% & 73.5\% & 84.0\% \\
Test land pixel-days & 11.6M & 102.0M & 31.8M \\
Precip.\ regime & Convective & Atm.\ rivers & TC + convective \\
Obs.\ mean (mm/day) & 4.14 & 1.29 & 3.98 \\
\bottomrule
\end{tabular}
\vspace{-0.5\baselineskip}
\end{wraptable}

We downscale daily ERA5 reanalysis at ${\sim}25$\,km resolution to PRISM
precipitation at ${\sim}4$\,km. Let $x \in \mathbb{R}^{C \times H \times W}$ be
the coarse input over $C$=15 ERA5 channels: 2-meter temperature with daily
extrema and soil temperature (\texttt{t2m}, \texttt{mx2t}, \texttt{mn2t},
\texttt{stl1}); 10-meter wind (\texttt{u10}, \texttt{v10}); precipitation totals
with daily extremes (\texttt{tp}, \texttt{cp}, \texttt{mxtpr}, \texttt{mntpr});
cloud cover at three levels (\texttt{lcc}, \texttt{tcc}, \texttt{hcc}); and
surface runoff and evaporation (\texttt{sro}, \texttt{e}). Let
$y \in \mathbb{R}_{\ge 0}^{H' \times W'}$ be the fine-resolution PRISM target.
A land mask $m \in \{0,1\}^{H' \times W'}$ restricts loss and evaluation to the
$N_\text{land} = \sum_{h,w} m_{h,w}$ land pixels. A deterministic downscaler
returns a single point estimate $\hat{y}_{h,w}$; we instead predict a discrete
conditional CDF comprising four quantile estimates $\hat{q}_{\tau,h,w}$ per
pixel for $\tau \in \mathcal{T} = \{0.50, 0.95, 0.99, 0.999\}$.
Table~\ref{tab:domains} summarizes the three study domains, which share the
same input variables and training period but differ sharply in
extreme-precipitation regime.

\subsection{SRDRN Backbone}
\label{subsec:backbone}

All models in this paper share a common feature extractor: the \srdrn{} of
\citet{wang2021deep}, a convolutional super-resolution network that bilinearly
upsamples $x$ to the PRISM grid, refines it through 16 residual blocks with
SpatialDropout2D, and recovers fine-scale detail through asymmetric pixel-shuffle
upsampling; we call the resulting per-pixel feature map the \emph{shared backbone
features}. The deterministic baseline maps these features to $\hat{y}$ through a
single $9 \times 9$ \texttt{Conv2D(1)} head trained with intensity-weighted MAE,
$w = \mathrm{clip}(y/11.7,\; 0.1,\; 2.0)$, where $11.7$\,mm/day is the
training-set standard deviation of wet-day precipitation and the weight saturates
at $y = 23.4$\,mm/day. \qsrdrn{} reuses this backbone unchanged, replacing only
the output head and the loss; any performance difference is therefore attributable
to the quantile design, not to feature extraction.

\subsection{Q-SRDRN: Multi-Quantile Regression}
\label{subsec:qsrdrn}

Turning the deterministic head into a multi-quantile predictor exposes four
coupled problems: \textbf{(1)} an asymmetric per-quantile loss; \textbf{(2)}
sufficient gradient signal at the upper tail despite ${\sim}1\%$ extreme-day
prevalence; \textbf{(3)} non-crossing monotonicity in $\tau$ that does not
destroy spatial filter specialization in a CNN; and \textbf{(4)} per-channel
gradient routing so the median and P999 detectors do not compete.

{\bf (1) Pinball loss.}
For quantile $\tau$ and residual $e = y - \hat{q}_\tau$, the pinball
loss~\citep{koenker2001quantile, bremnes2004probabilistic}
$\rho_\tau(e) = \max(\tau e, (\tau - 1) e)$
penalizes under-prediction $\tau/(1-\tau)$ times more heavily than over-prediction
($99{\times}$ at $\tau{=}0.99$).  Aggregated over $\mathcal{T} = \{0.50,
0.95, 0.99, 0.999\}$ and land pixels:
\begin{equation}
  \mathcal{L} = \sum_{\tau \in \mathcal{T}} \frac{1}{N_\text{land}}
  \sum_{h,w} m_{h,w}\, w_\tau\, \rho_\tau\!\bigl(y_{h,w} - \hat{q}_{\tau,h,w}\bigr).
\label{eq:loss}
\end{equation}

{\bf (2) Per-quantile event weighting.}
$w_\tau = 1 + \alpha\, \mathbf{1}[y > z_\text{thresh}]$ with $\alpha = 5$ on the
upper-tail heads (P95/P99/P999) upweights extreme days by $6{\times}$;
$z_\text{thresh}$ is the 75th-percentile precipitation in log1p z-score units.  The
P50 head is exempt ($w_{0.50}{=}1$) so its convergence to the true conditional
median---the property that lets it absorb augmentation without bias
(\S\ref{subsec:cvae})---is preserved.

{\bf (3) IncrementBound: gradient-stable monotonicity.}
The textbook non-crossing fix---per-pixel sorting of output channels---is
destructive on a CNN: sorting induces a per-pixel permutation, so the same
convolutional filter receives P999-scale gradients at some pixels and
P50-scale gradients at others, preventing any filter from specializing. Adapting the
cumulative-softplus construction of \citet{cannon2018non} pins each output
channel to a fixed quantile identity:
\begin{equation}
  \hat{q}_{0.50} = 8\tanh(r_0 / 8), \quad
  \hat{q}_{\tau_{k}} = \hat{q}_{\tau_{k-1}} + \mathrm{softplus}(r_k), \quad k = 1,2,3.
\label{eq:increment}
\end{equation}
This gives $\hat{q}_{0.50} \le \hat{q}_{0.95} \le \hat{q}_{0.99} \le \hat{q}_{0.999}$
by construction, fixes the $k$-th channel's gradient identity, and lets the
upper quantiles grow without an explicit cap (a physical $1{,}200$\,mm cap
is applied only in post-processing).

{\bf (4) Separate output heads.}
A shared \texttt{Conv2D(4)} output layer forces all four quantiles into the
same filter bank, but P50 ($\approx 0$ at most pixels) and P999 (rare extreme
firings) need fundamentally different spatial detectors.  We use four
independent \texttt{Conv2D(1)} heads (one $9{\times}9$ kernel each), concatenated
and passed through IncrementBound, so every quantile gets its own dedicated
spatial detector.

\begin{figure}[t]
  \centering
  \includegraphics[width=\textwidth]{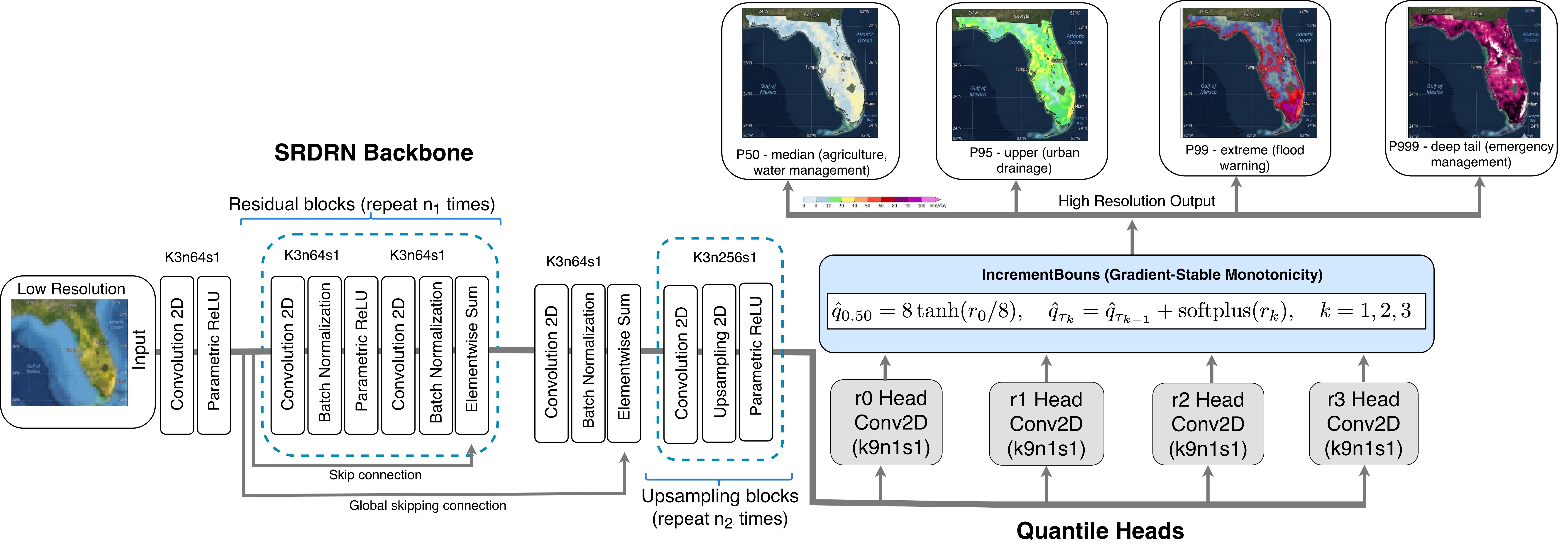}
  \caption{\textbf{Q-SRDRN architecture.}  The shared backbone (16 residual blocks)
  feeds four separate Conv2D(1) output heads.  IncrementBound enforces
  monotonicity via cumulative softplus increments (Eq.~\ref{eq:increment}).
  P50 receives pure pinball loss; P95/P99/P999 receive event-weighted pinball loss
  ($\alpha=5$).}
  \label{fig:architecture}
\end{figure}

\subsection{cVAE Augmentation}
\label{subsec:cvae}

In regimes where extreme spatial patterns are sparse in the historical record,
the P50 head is data-starved at the thresholds that matter most.  Synthetic
augmentation is the natural remedy, and---unlike under MAE---works here
because the upper-tail heads no longer rely on the median to reach extremes,
so the median can absorb augmented samples without contaminating them.  We
use a conditional VAE~\citep{sohn2015learning}, building on the variational
autoencoder framework of~\citet{kingma2014auto}, that learns a
stochastic ERA5\,$\to$\,PRISM mapping with a $256$-dim Gaussian latent
$z$; skip connections are omitted (forcing the decoder to use $z$), GroupNorm
is used for train/eval consistency, and $\beta_\text{KL}{=}0.008$ keeps the
posterior informative.  For large output grids the fully-connected decoder
becomes a bottleneck (e.g., California's spatial correlation is $0.707$ on $76$K pixels), so we
add a \emph{Spatial \cvae{}} variant that removes the FC layer and upsamples
through convolutional blocks with AdaIN $z$-conditioning, raising California's
correlation to $0.835$.  Florida uses $83$ synthetic samples (${\sim}0.67\%$
of training; the cliff to harm is near ${\sim}1\%$); California uses $385$
samples (${\sim}3.1\%$, tolerable because the upper-tail channel is augmentation-independent
in this regime).  Full architecture, training, and augmentation
budgets are in Appendix~\ref{app:augmentation}.
\section{Experiments}
\label{sec:experiments}

\subsection{Experimental Design}

Our experiments are organized as a single $2 \times 2$ factorial study
that cleanly separates the architecture contribution from the augmentation
contribution.  We evaluate four model configurations per domain---the cross of
\{\srdrn{}, \qsrdrn{}\} with \{no augmentation, \cvae{} augmentation\}---so that
every comparison either fixes the loss family and varies the data, or fixes the
data and varies the loss family.  All models share the same backbone, ERA5
inputs, and training/test splits.  \qsrdrn{} additionally uses separate output
heads, IncrementBound, and per-quantile event weighting ($\alpha=5$).  Test-set
metrics are reported throughout.

At each precipitation threshold $T$ we form the $2{\times}2$ contingency table
with $a$\,=\,hits ($\hat y > T$ and $y > T$), $b$\,=\,false alarms
($\hat y > T$ and $y \le T$), $c$\,=\,misses ($\hat y \le T$ and $y > T$), and
$d$\,=\,correct rejections.  The \textbf{Probability of Detection}, also called the hit rate,
POD\,$=a/(a+c)$, is the share of observed extreme
events the model captures, and the \textbf{False Alarm Ratio},
FAR\,$=b/(a+b)$, is the share of forecast events that do not occur.  Neither
score is informative on its own at extreme thresholds---``always wet'' achieves
POD\,$=$\,1 and ``always dry'' achieves FAR\,$=$\,0, both with zero skill---so
we report them only where the absolute event count is itself informative.  The
\textbf{Symmetric Extremal Dependence Index}
(SEDI,~\citealp{ferro2011extremal}) combines POD and the false-alarm rate
$b/(b+d)$ into a single skill score that converges to a non-degenerate limit as
the base rate $\to$\,0, making it our headline extreme-event metric.  Kullback--Leibler (KL)
divergence measures distributional fidelity, while Continuous Ranked Probability Score (CRPS) and interval coverage (available only for \qsrdrn{}) assess calibration quality.

\subsection{Architecture: \srdrn{} vs.\ \qsrdrn{} across three domains}
\label{subsec:architecture}

\begin{wraptable}[18]{r}{0.39\textwidth}
\vspace{-0.6\baselineskip}
\centering
\caption{\textbf{Bulk fit (central-tendency channels).}
P95, P99, and P999 heads target levels above the observation by
construction.
\best{Bold} is best in row.}
\label{tab:bulk}
\vspace{2pt}
\footnotesize
\setlength{\tabcolsep}{4pt}
\renewcommand{\arraystretch}{1.05}
\begin{tabular}{lccc}
\toprule
 & \textbf{RMSE} & \textbf{$r$} & \textbf{KL} \\
\midrule
\multicolumn{4}{l}{\textit{Florida}} \\
\srdrn{}        & 9.162  & 0.599  & 0.124  \\
\qsrdrn{} P50   & \best{8.804}  & \best{0.631}  & \best{0.046}  \\
\midrule
\multicolumn{4}{l}{\textit{California}} \\
\srdrn{}        & 2.944  & 0.843  & 0.0060 \\
\qsrdrn{} P50   & \best{2.897}  & \best{0.845}  & \best{0.0032} \\
\midrule
\multicolumn{4}{l}{\textit{Texas substate}} \\
\srdrn{}        & 10.589 & 0.624  & 0.176  \\
\qsrdrn{} P50   & \best{9.461}  & \best{0.698}  & \best{0.014}  \\
\bottomrule
\end{tabular}
\end{wraptable}

We begin with the architecture-only comparison: \srdrn{}'s single output
versus the four heads of \qsrdrn{}, both trained on the original (un-augmented)
ERA5/PRISM data, on Florida, California, and the Texas substate.  This
isolates what the multi-head pinball architecture buys before any augmentation
is introduced; \cvae{}-augmented results, which apply only to Florida, are
deferred to \S\ref{subsec:augmentation}.  Bulk fit (RMSE, Pearson $r$,
KL divergence) is summarised in Table~\ref{tab:bulk} for the central-tendency
predictors (\srdrn{} and the \qsrdrn{} P50 head); upper-quantile heads
(P95/P99/P999) target distributional levels above the observation by design,
so RMSE/Pearson/KL are not informative for them and we score them through
their natural extreme-detection metrics.  Table~\ref{tab:arch} reports SEDI
and hit counts across all four \qsrdrn{} heads at thresholds matched to each
head's climatological exceedance level (P95 at 10--20\,mm, P99 at 30--60\,mm,
P999 at $\ge$70\,mm).  P50 and matched-head SEDI envelopes are visualised in
Appendex~\ref{app:p50_sedi_fig} and~\ref{app:matched_head_fig}.

\begin{table}[!htbp]
\centering
\caption{\textbf{Architecture comparison (without no augmentation).}
\srdrn{} (single output) vs.\ \qsrdrn{}.
\best{Bold}\,=\,best per-threshold within each domain. 
}
\label{tab:arch}
\vspace{4pt}
\footnotesize
\renewcommand{\arraystretch}{1.05}
\setlength{\tabcolsep}{4pt}
\begin{tabular}{lccccccc}
\toprule
 & \multicolumn{5}{c}{\textbf{SEDI at threshold}} & \multicolumn{2}{c}{\textbf{Hits}} \\
\cmidrule(lr){2-6} \cmidrule(lr){7-8}
\textbf{Model} & \textbf{50} & \textbf{100} & \textbf{150} & \textbf{200} & \textbf{300} & \textbf{@200\,mm} & \textbf{@300\,mm} \\
\midrule
\multicolumn{8}{l}{\textit{Florida} \;($n_{200}{=}2{,}111$, $n_{300}{=}75$)} \\
\srdrn{}              & 0.572 & 0.694 & 0.725 & 0.564 & 0.000 & 88    & 0 \\
\qsrdrn{} P50         & 0.522 & 0.688 & 0.607 & 0.405 & 0.000 & 14    & 0 \\
\qsrdrn{} P95         & 0.712 & 0.739 & 0.831 & 0.831 & 0.000 & 899   & 0 \\
\qsrdrn{} P99         & \best{0.778} & 0.799 & 0.852 & 0.896 & 0.352 & 1{,}309 & 1 \\
\qsrdrn{} P999        & 0.774 & \best{0.859} & \best{0.879} & \best{0.932} & \best{0.710} & \best{1{,}598} & \best{19} \\
\midrule
\multicolumn{8}{l}{\textit{California} \;($n_{200}{=}537$, $n_{300}{=}15$)} \\
\srdrn{}              & 0.823 & 0.777 & 0.755 & 0.705 & 0.000 & 54    & 0 \\
\qsrdrn{} P50         & 0.826 & 0.824 & 0.825 & 0.833 & 0.794 & 179   & 3 \\
\qsrdrn{} P95         & 0.972 & 0.975 & 0.983 & 0.980 & \best{1.000} & 484 & \best{15} \\
\qsrdrn{} P99         & 0.988 & 0.992 & 0.994 & 0.994 & 1.000 & 521   & 15 \\
\qsrdrn{} P999        & \best{0.989} & \best{0.996} & \best{0.997} & \best{0.998} & 1.000 & \best{534} & 15 \\
\midrule
\multicolumn{8}{l}{\textit{Texas substate} \;($n_{200}{=}10{,}720$, $n_{300}{=}2{,}265$)} \\
\srdrn{}              & 0.609 & 0.550 & 0.462 & 0.240 & 0.000 & 2     & 0 \\
\qsrdrn{} P50         & 0.601 & 0.604 & 0.534 & 0.263 & 0.000 & 7     & 0 \\
\qsrdrn{} P95         & 0.847 & 0.815 & 0.784 & 0.794 & 0.558 & 4{,}143 & 160 \\
\qsrdrn{} P99         & \best{0.899} & 0.882 & 0.871 & 0.868 & 0.758 & 6{,}307 & 695 \\
\qsrdrn{} P999        & 0.836 & \best{0.935} & \best{0.934} & \best{0.942} & \best{0.891} & \best{8{,}776} & \best{1{,}441} \\
\bottomrule
\end{tabular}
\vspace{-1em}
\end{table}

\textbf{Bulk performance.} \qsrdrn{}'s P50 channel attains lower RMSE, higher Pearson $r$, and lower KL
divergence than \srdrn{} on all three domains (Table~\ref{tab:bulk}).  The
largest improvement is on the heaviest-tailed regime (Texas substate:
RMSE $-10.6\%$, KL $0.176 \to 0.014$); on California, where \srdrn{} is
already well-calibrated by virtue of the atmospheric-river regime, the gain
is smaller (RMSE $-1.6\%$).  The diagnostic is sharpest on
Florida, where \srdrn{}'s 43\% over-prediction bias (predicted mean 5.94
vs.\ observed 4.14\,mm/day) inflates extreme-threshold hits at the cost of
distributional fidelity (KL 0.124 vs.\ 0.046).

\textbf{P50 extreme detection.} The three domains tell different stories at the extreme tail.  On California,
the P50 head alone already dominates \srdrn{} at every threshold above
100\,mm: SEDI at 200\,mm rises from 0.705 to 0.833 ($+18\%$), and at
300\,mm---where \srdrn{} is blind---P50 reaches 0.794.

On Florida and the Texas substate the comparison reveals a calibration--detection
trade-off.  \qsrdrn{}'s P50 without augmentation is more conservative at extreme
thresholds than \srdrn{} (FL SEDI at 200\,mm: 0.405 vs.\ 0.564), because the true
conditional median is low for most pixels while \srdrn{}'s bulk over-prediction
inflates hits at the cost of distributional fidelity (KL 0.124 vs.\ 0.046).  In
short, \srdrn{} buys SEDI at extreme thresholds with bulk over-prediction, whereas
\qsrdrn{}'s P50 buys distributional calibration at the cost of extreme-day
exposure.  The multi-head design resolves the trade-off without choosing: each
remaining head is the optimal detector for the threshold range matching its
climatological exceedance level, and the matched-head envelope dominates
\srdrn{} at every threshold on all three domains (Table~\ref{tab:arch},
P95--P999 rows; matched-head envelope in Appendix~\ref{app:matched_head_fig}).

\textbf{P999 detection across the three domains.} The P999 head delivers the headline lift on every domain.  On Florida it
raises POD at 200\,mm from $4.2\%$ ($88$ hits) to $75.7\%$ ($1{,}598$ hits) and
detects $19$ of $75$ events at $300$\,mm where \srdrn{} is blind
(SEDI\,$=$\,$0.710$).  On California it reaches near-perfect detection
(SEDI\,$\ge$\,$0.996$ through $300$\,mm; $534$ of $537$ events at $200$\,mm;
$15$ of $15$ at $300$\,mm), reflecting California's atmospheric-river regime
which provides a clearer large-scale signal than Florida's scattered
convection.  The Texas substate is the most extreme demonstration: \srdrn{}
catches $2$ of $10{,}720$ events at $200$\,mm and zero of $2{,}265$ at
$300$\,mm, while the \qsrdrn{} P999 head catches $8{,}776$
(POD\,$=$\,$81.9\%$, SEDI\,$=$\,$0.942$) and $1{,}441$
(POD\,$=$\,$63.6\%$, SEDI\,$=$\,$0.891$) respectively---a regime with
$\sim$5$\times$ more 200\,mm events than Florida and $\sim$30$\times$ more at
$300$\,mm, so the deep-tail detection signal is statistically resolved in a
way Florida and California cannot match.

\textbf{Head selection provides the operational advantage.} Beyond raw skill numbers, the multi-head design lets operational users select
the quantile head matching their risk tolerance: P50 (agriculture/water
management), P95 (urban drainage), P99 (flood warning), P999 (emergency
management).  The spread between P50 and P999 also encodes tail
risk directly: two pixels with identical 10\,mm at P50 would carry vastly different flood
risk for 20 mm vs.\ 220\,mm at P999. Such information is invisible to any deterministic model.

\subsection{IncrementBound and Separate Heads}
\label{subsec:ablation}

Having established the gain from specialized heads, we next isolate the two
architectural choices that make that specialization possible.  Both target
a single failure mode: the loss of gradient identity in spatial quantile
heads. In this section, we discuss the diagnosis, the fix, and a capacity-control
ablation that rules out the most natural alternative explanation.

\textbf{Problem:} Multi-quantile regression requires that predicted quantiles never cross:
$\hat{q}_{0.5} \le \hat{q}_{0.95} \le \hat{q}_{0.99} \le \hat{q}_{0.999}$ for every pixel.
Without enforcement, a CNN can freely predict $\hat{q}_{0.5} > \hat{q}_{0.99}$---a
mathematically invalid output that violates the definition of a quantile function
($F^{-1}(\tau)$ is non-decreasing in $\tau$).
The standard fix is to sort the output channels: $\hat{\mathbf{q}} = \mathrm{sort}(\mathbf{r})$,
where $\mathbf{r}$ are the raw network outputs.
Sorting guarantees valid quantiles but introduces a permutation that potentially changes at every pixel.
During backpropagation, the P999 pinball gradient flows to whichever raw channel
happened to be largest---a different channel at each pixel and at each forward
pass. Thus, no channel can specialize: the P999 head receives P50-scale gradients
half the time and vice versa.  
Enforcing quantile monotonicity destroys gradient identity.

\textbf{Solution:} We replace \texttt{tf.sort} with the cumulative-softplus parameterisation of
Eq.~\ref{eq:increment}, which guarantees monotonicity by construction while
preserving a fixed gradient identity for each quantile---the P999 channel
always receives P999 gradients.  Combined with four separate Conv2D(1) output
heads (eliminating filter competition in a shared output layer), this yields
significant improvement.

A capacity-control ablation (Appendix~\ref{app:ablation}) rules out the
alternative hypothesis that P999 detection is parameter-limited: adding a
$259$K-parameter \texttt{Conv2D(64)+PReLU} expansion before the P999 output
collapses SEDI@300\,mm from $0.735$ to $-0.210$ (zero hits) and P50 SEDI@200\,mm
from $0.866$ to $0.000$.  The $300$\,mm ceiling is set by the $25$\,km ERA5
input resolution, not by model capacity.

\subsection{Comparison Against Uncertainty-Quantification Baselines}
\label{subsec:uq_baselines}

To verify that \qsrdrn{}'s gains are not reproducible by simpler probabilistic
schemes, we ablate against four external baselines that fix the backbone,
training data, and 83-sample \cvae{} augmentation and vary only the
predictive head:
\textbf{(i) \textsc{mc\_dropout}}~\citep{gal2016dropout} extracts empirical
P50/P95/P99/P999 from $K{=}30$ stochastic forward passes; the predictive
variance turns out to be approximately quantile-invariant, so all upper
quantiles collapse onto roughly the same wet-day band, giving an empirical
P999 exceedance of $25.5\%$ on Florida ($250{\times}$ above the $0.10\%$
target) and $14.4\%$ on California ($144{\times}$).
\textbf{(ii) \textsc{single\_head\_qr}}~\citep{cannon2011quantile} is a
shared-head QR-CNN in the standard QRNN lineage, with post-hoc
\texttt{np.sort} non-crossing applied at inference; it matches \qsrdrn{} on
bulk fit and on deep-tail P999 detection---the cost of shared-head sorting
is paid in the predictive-\emph{interval shape}, not in bulk skill or event
hits. The gradient-permutation pathology of \S\ref{subsec:ablation}
 widens the P99$-$P50 spread to $38$--$138$\,mm vs.\
$8$--$22$\,mm for \qsrdrn{}+\cvae{} and inflates KL by $41\%$ ($0.032$ vs.\
$0.023$), so the predicted quantiles cannot be used as calibrated decision
intervals.
\textbf{(iii) \textsc{deep\_ensemble}}~\citep{lakshminarayanan2017simple}
quantile-aggregates $M{=}5$ deterministic SRDRNs (rather than the dual-head
Gaussian-NLL form of the original paper), so the ensemble spread captures only
epistemic model uncertainty and not heteroscedastic data noise; it consequently
reproduces the same quantile-invariant variance as \textsc{mc\_dropout} via a
different mechanism, with $25.0\%$ empirical P999 exceedance on Florida.
\textbf{(iv) \textsc{multi\_seed}} runs five seeds of \qsrdrn{} itself
($\{11,23,47,71,97\}$); all five land within a tight band (RMSE
$8.640\,\pm\,0.042$, P999 SEDI@200\,mm $0.928\,\pm\,0.004$) confirming
that the headline numbers are not a lucky seed.

\textbf{Sampling-based tail failure is structural, not a tuning artifact.}
\textsc{mc\_dropout}'s weak Florida bulk fit (RMSE $16.7$ vs.\ \qsrdrn{}'s
$8.7$, Pearson $0.33$ vs.\ $0.63$) has a clear mechanistic origin rather
than a tuning origin: applying dropout inside a CNN acts as an aggressive
spatial regularizer that smears Florida's localized convective cells, which
degrades any pixel-wise mean-field metric. Two independent comparisons
confirm this attribution. First, the same \textsc{mc\_dropout}
configuration is essentially bulk-competitive with \qsrdrn{} on
California's smoother synoptic/orographic precipitation (RMSE $2.99$ vs.\
$2.88$, Pearson $0.84$ vs.\ $0.85$), so the Florida bulk gap is a
climatology-dependent regularization effect, not a hyperparameter failure.
Second, \textsc{deep\_ensemble} aggregates deterministic SRDRNs
\emph{without} internal dropout and therefore recovers \qsrdrn{}'s bulk
RMSE and Pearson everywhere ($8.62$/$0.65$ on Florida, $2.73$/$0.86$ on
California). Yet whether the bulk fit is poor (\textsc{mc\_dropout}
Florida), competitive (\textsc{mc\_dropout} California), or matched
(\textsc{deep\_ensemble} both regions), all three settings collapse to
the \emph{same} $\tau$-invariant predictive variance and a near-identical
$\sim\!25\%$ Florida and $\sim\!13$--$14\%$ California P999 exceedance ---
$144$--$250\times$ above the $0.10\%$ target.  The inability to capture
heavy-tail precipitation is therefore a structural property of
sampling-based UQ in deep learning---epistemic spread alone cannot
stretch to cover heavy-tailed targets---and not an artifact of either
spatial blurring or sub-optimal baseline configuration.

Full table and per-baseline
diagnostics, including CA subset and multi-seed
calibration check, are in Appendix~\ref{app:uq_baselines}.  In one sentence:
explicit per-quantile pinball regression directly optimizes the target loss
at each $\tau$ and is not substitutable by post-hoc empirical-quantile
wrappers around point predictors.

\subsection{\cvae{} Augmentation}
\label{subsec:augmentation}

A region-specific \cvae{} \emph{generator} was trained on each of the three
domains, but the \cvae{} \emph{architecture and input-variable set} were
tuned only for Florida and reused unchanged on California and the Texas substate.
Using this configuration, augmentation \emph{helped} Florida
but \emph{mildly harmed} California's P50 channel (P50 SEDI@200\,mm
$0.833\to0.739$). We only present results for Florida
(Table~\ref{tab:aug}).

\begin{table}[t]
\centering
\caption{\textbf{\cvae{} augmentation on Florida.}
The 83-sample \cvae{} augmentation set is the same one used by the headline
\qsrdrn{}+\cvae{} runs throughout the paper.  \best{Bold}\,=\,best in row.
P999 rows are reported only for \qsrdrn{} configurations because \srdrn{}
is single-output.  California and Texas-substate \cvae{}-augmented entries
are omitted; see the discussion above for why.}
\label{tab:aug}
\vspace{4pt}
\footnotesize
\renewcommand{\arraystretch}{1.05}
\begin{tabular}{lcccc}
\toprule
\textbf{Metric} & \textbf{\srdrn{}} & \textbf{\srdrn{}+\cvae{}} & \textbf{\qsrdrn{}} & \textbf{\qsrdrn{}+\cvae{}} \\
\midrule
RMSE (mm/day)             & 9.162 & 8.973 & 8.804 & \best{8.665} \\
Pearson $r$               & 0.599 & 0.611 & \best{0.631} & 0.629 \\
KL divergence             & 0.124 & 0.126 & 0.046 & \best{0.023} \\
\addlinespace
P50 SEDI @ 200\,mm        & 0.564 & 0.566 & 0.405 & \best{0.866} \\
P50 SEDI @ 300\,mm        & 0.000 & 0.000 & 0.000 & \best{0.555} \\
P50 hits @ 200\,mm ($n{=}2{,}111$) & 88 & 95 & 14 & \best{1{,}038} \\
\addlinespace
P999 SEDI @ 200\,mm       & ---   & ---   & \best{0.932} & 0.922 \\
P999 SEDI @ 300\,mm       & ---   & ---   & 0.710 & \best{0.735} \\
P999 hits @ 200\,mm       & ---   & ---   & \best{1{,}598} & 1{,}558 \\
P999 hits @ 300\,mm ($n{=}75$) & --- & --- & 19 & \best{23} \\
\bottomrule
\end{tabular}
\vspace{-1.8em}
\end{table}

\textbf{Bulk and P50 detection.} \qsrdrn{}+\cvae{} attains the best RMSE ($-5.4\%$) and KL ($-81\%$);
augmentation lifts the P50 head from 14 to $1{,}038$ hits at 200\,mm
($74{\times}$) and SEDI@200\,mm from 0.405 to 0.866, while the matched-head
P999 channel is essentially unchanged (SEDI@200\,mm $0.932\to0.922$,
hits@300\,mm $19\to23$): augmentation concentrates its value in the median
channel.  The asymmetry against \srdrn{} follows from the loss: MAE amplifies
label conflicts ($20{\times}$ gradient on a synthetic 170\,mm pixel),
whereas pinball treats real and synthetic pixels equally, so 83 synthetic
days among $12{,}400$ shift the median by $<$$0.6\%$ (full mechanism in
Appendix~\ref{app:aug_mechanism}).

\textbf{Augmentation is region-specific.} The lesson is that the augmentation architecture and input-channel
set are themselves regime dependent: convective (FL), atmospheric-river (CA),
and tropical-cyclone-coastal (Texas substate) regimes likely require different
generator backbones, latent-dim budgets, and conditioning variables.  Multi-quantile
regression, by contrast, transfers across all three demains without retuning. 

\section{Discussion}
\label{sec:discussion}

Florida and California are not independent benchmarks; they are two halves of
the same contribution probed from opposite directions, with the Texas substate
adding a third architecture-only confirmation. The remainder of this section
reads the empirical record through that decomposition: the relative roles of
architecture and augmentation across the three regimes, the quality of the
predictive distribution itself, and the limits beyond which the framework
cannot extend.

\subsection{Two Domains, Two Roles for the Two Components}
\label{subsec:decomposition}

The three domains isolate the two halves of the contribution.  California
(atmospheric-river regime) is the architecture-only test bed: un-augmented
\qsrdrn{} already reaches P999 SEDI\,$\ge$\,0.99 at every threshold through
$300$\,mm/day and P50 SEDI\,$=$\,0.794 at $300$\,mm.  Florida (sparse
scattered convection) is the rescue test bed: extreme spatial patterns are
too rare for the median head to learn from real data alone, and 83 synthetic
\cvae{} samples ($\sim$0.67\% of training) lift P50 SEDI@200\,mm from $0.405$
to $0.866$.  The Texas substate is a third architecture-only confirmation
on a heavier-tailed regime ($n_{200}{=}10{,}720$, $n_{300}{=}2{,}265$).  The
full per-component decomposition is in Appendix~\ref{app:contributions}
(Table~\ref{tab:contributions}); the pattern is consistent across all three
domains, with the CA P50 entry telling the most informative story:
augmentation \emph{decreases} P50 SEDI@200\,mm by $0.094$ because synthetic
samples smooth the conditional median toward the population mean in a regime
that did not need them, while seasonal RMSE (Table~\ref{tab:seasonal} in
Appendix~\ref{app:seasonal}) reinforces this---\qsrdrn{}+\cvae{} wins seasonally
on FL but only sporadically on CA.  We therefore recommend \qsrdrn{} (no
augmentation) as the operational California and Texas-substate model and
\qsrdrn{}+\cvae{} as the operational Florida model: the two components are
complementary, not additive.

\subsection{Probabilistic Calibration Diagnostics}
\label{subsec:calibration}

Beyond point skill, the predictive distribution itself is calibrated: all
eight head$\times$domain combinations show positive calibration-gap closure
($5$--$24\%$, mean $13\%$); FL $[\text{P50},\text{P99}]$ interval coverage
is $46.9\%$ (target $49\%$) and CA is $48.0\%$.  CRPS skill by intensity
band mirrors the regime split: \cvae{} improves CRPS in every FL band
$\ge$10\,mm (peak $+17.7\%$ at $\ge$200\,mm) but flips to $-29.5\%$ on
CA's heaviest band.  Sharpness pinpoints the mechanism.  CRPS combines a
sharpness term and a calibration term, and the CA loss is in calibration:
\cvae{} narrows the predictive spread on \emph{both} domains' extreme
bands (sharpness improves), but on CA the narrower distribution
concentrates around a biased median (P50 SEDI $-0.094$,
Table~\ref{tab:contributions}; cf.\ \S\ref{subsec:decomposition})---overconfidence,
not skill.  Full
reliability, CRPS, and sharpness diagrams are in
Appendix~\ref{app:calibration}.

\subsection{Limitations}
\label{subsec:limitations}

\textbf{ERA5 resolution wall.} At $300$\,mm/day only $75$ events exist in
Florida's $11.6$M pixel-day test set, and a capacity-expansion ablation
(Appendix~\ref{app:ablation}) collapses detection entirely---the $25$\,km ERA5
forcing cannot resolve the meso-$\gamma$-scale dynamics that produce
ultra-extreme point rainfall.  Higher-resolution forcing (HRRR $3$\,km, $1$\,km
RCM) is the natural next step; the architecture is input-resolution agnostic.

\textbf{Further limits.} \cvae{} generator smoothness vs.\ diffusion, residual
P999 calibration gap ($15.12{\times}$ wet-day on FL), and transfer beyond
precipitation are detailed in Appendix~\ref{app:limitations_extra}. Headline
numbers are single-seed; FL seed variance is bounded by the 5-seed
\textsc{multi\_seed} comparator (Appendix~\ref{app:uq_baselines}); the CA
\textsc{multi\_seed} run lands in the camera-ready.

\subsection{Broader Impacts and Reproducibility}
\label{subsec:broader-impacts}

\textit{Broader impacts.} Improved downscaling supports flood-risk, water,
agricultural, and climate-adaptation planning in hurricane- and
atmospheric-river-prone regions; risks are limited to over-trusting the
median or applying the model outside training domains/period
(1980--2013 ERA5--PRISM)---the calibrated quantile heads, not a point
prediction, are the intended interface.
\textit{Reproducibility.} ERA5 and PRISM are public; full code (training
scripts, pinball+IncrementBound, postprocessing, \cvae{} pipeline, figure scripts) is released on acceptance.  Each \qsrdrn{}
run is 12\,h (FL and TX) / 22\,h (CA) on one H100 (80\,GB);
hyperparameters in Appendix~\ref{app:hyperparams}.

\section{Conclusion}
\label{sec:conclusion}

MAE-trained downscalers fail at extremes because their conditional median
cannot reach the upper tail.  Replacing the loss with multi-quantile pinball
regression, gradient-decoupling the heads via IncrementBound and separate
output convolutions, and---only where the regime requires it---adding a
small \cvae{}-generated extreme set, lifts P999 detection at 200\,mm/day
from $88\to 1{,}598$ events on FL ($18{\times}$), to SEDI\,$\ge$\,0.996
through 300\,mm on CA, and from 2 of $10{,}720$ to $8{,}776$ ($81.9\%$) on
the Texas substate, all without scaling up the network.  The dominant
bottleneck at extremes is the loss, not the network or the regime.

\begin{ack}
Omitted for double-blind review.
\end{ack}


\bibliographystyle{plainnat}


\newpage
\appendix

\section{Architectural Ablation Details}
\label{app:ablation}

We add architectural changes one at a time, evaluating each step against the
previous configuration. \textbf{Step~(i)} is the baseline: \qsrdrn{} with
\texttt{tf.sort} for monotonicity and a shared \texttt{Conv2D(4)} output layer.
Steps~(ii)--(iv) below progressively replace these components, each in
isolation, with backbone, loss, data, and epochs held identical across the
chain.

\paragraph{Step~(ii): IncrementBound.}
We replace \texttt{tf.sort} with the cumulative-softplus parameterisation of
Eq.~\ref{eq:increment}. The prior tanh bound on $r_0$ becomes unnecessary
because monotonicity is now guaranteed by construction; upper quantiles grow
without an explicit cap (physical caps of $1{,}200$\,mm are applied only in
post-processing).

P999 detection improves at all thresholds $\ge200$\,mm: POD at $300$\,mm nearly
doubles ($0.133 \to 0.253$) and SEDI rises from $0.567$ to $0.708$ ($+24.9\%$).
A side effect: P50 SEDI at $150$--$250$\,mm also improves (e.g., $+22\%$ at
$250$\,mm), because consistent gradient routing lets each channel's filters
specialize rather than averaging the four quantile signals.

\paragraph{Step~(iii): Separate output heads.}
We replace the shared \texttt{Conv2D(4,\,kernel=9)} with four independent
\texttt{Conv2D(1,\,kernel=9)} heads, preserving total parameter count. P999
SEDI at $300$\,mm improves further to $0.735$ ($+3.8\%$); P50 SEDI at
$200$\,mm jumps from $0.684$ to $0.866$ ($+26.6\%$); KL divergence reaches
its best value of $0.023$.

\paragraph{Step~(iv): Deeper P999 head (FAILED).}
We add a \texttt{Conv2D(64,\,kernel=3)+PReLU} expansion before the P999
output head ($259$K added parameters). P999 SEDI at $300$\,mm collapses from
$0.735$ to $-0.210$ (zero hits). The failure has three coupled causes: the
$512 \to 64$ channel reduction loses spatial information, the $259$K extra
parameters trained on ${\sim}0.1\%$ of pixel-days learn a conservative
mapping, and the degraded P999 gradients propagate back through IncrementBound
into the shared backbone. The result confirms that the $300$\,mm detection
limit is imposed by ERA5's $25$\,km input resolution, not by model capacity.

\section{Augmentation Experiments Summary}
\label{app:augmentation}

Across 23 \srdrn{} augmentation experiments, only one configuration (83 samples, 0.67\%
ratio) succeeded.  Key failures include:
\begin{itemize}[noitemsep]
  \item \textbf{Higher ratios}: 2.1\% (Corr\,$-0.007$), 1.5\%
    (RMSE\,$+0.16$, SEDI@100\,$-0.08$), 9.4\% (catastrophic collapse).
  \item \textbf{Two-stage fine-tuning}: Fine-tunning the trained SRDRN on non-augmented data with augmented data.  RMSE\,=\,10.81; pred max\,=\,1{,}919\,mm.
    No real-data anchor causes uncontrolled gradient drift.
  \item \textbf{Loss modification}: changing $w_\text{max}$\,=\,2.0 to $w_\text{max}$\,=\,5.0 (pred mean\,$+72.5\%$).
    Amplifies gradients for all events, not just synthetic.
\end{itemize}

For \qsrdrn{}, the 0.67\% ceiling persists: increasing to 186 samples (1.5\%) regressed on
RMSE ($+0.16$), Corr ($-0.02$), KL ($+52\%$), and SEDI@100 ($-0.08$).
The transition is a cliff near ${\sim}$1\%, confirming that the label-averaging
effect is the root cause.

\section{Failed Experiments Table}
\label{app:failures}

This appendix compiles the negative results referenced throughout the main
text, each establishing a hard constraint of the framework. The rows in
Table~\ref{tab:failures} fall into three groups: augmentation-ratio
failures that locate the ${\sim}1\%$ cliff (rows 1, 4, 6); loss and
optimisation failures from amplifying tail gradients indiscriminately
(rows 2, 5, 7); and capacity-versus-input-resolution failures that show
the deep-tail ceiling is set by ERA5 rather than by model depth (rows 3,
8).

\begin{table}[h]
\centering
\caption{Negative results establishing hard constraints of the framework.}
\label{tab:failures}
\small
\renewcommand{\arraystretch}{1.1}
\begin{tabular}{llp{6.5cm}}
\toprule
\textbf{Experiment} & \textbf{Failure mode} & \textbf{Diagnosis} \\
\midrule
SRDRN+266 samples & Corr $-$0.007 & Mild dilution at 2.1\% ratio \\
SRDRN, $w_\text{max}$=5 & pred mean $+$72.5\% & Amplified all gradients \\
SRDRN 2-stage fine-tune & RMSE 10.81, Corr 0.54 & No real-data anchor \\
SRDRN+1475 samples & Train RMSE worse & MAE averaged 3.7 + 178\,mm targets \\
Q-SRDRN, $\alpha$=2 uniform & P50 exceedance 25.6\% & Uniform $\alpha$ biases P50 at any $\alpha > 0$ \\
Q-SRDRN+186 samples & RMSE $+$0.16, KL $+$52\% & 1.5\% ratio; cliff near ${\sim}$1\% \\
Q-SRDRN, grad scaling & P50 unchanged & Zero-inflation, not gradient contamination \\
Q-SRDRN, deeper P999 & SEDI collapsed & 512$\to$64 bottleneck; ERA5 resolution limit \\
\bottomrule
\end{tabular}
\end{table}

\section{Why Augmentation Helps P50 but Not P999}
\label{app:aug_mechanism}

For quantile $\tau$, the pinball loss penalizes under-prediction $\tau/(1-\tau)$ times
more heavily than over-prediction.  This ratio determines how much extreme-event gradient
signal each head receives from the real training data alone:

\begin{center}
\small
\begin{tabular}{lccl}
\toprule
\textbf{Head} & $\boldsymbol{\tau}$ & \textbf{Under-pred.\ penalty} & \textbf{Augmentation effect} \\
\midrule
P50  & 0.50  & $1\times$ (symmetric) & \textbf{Essential} ($+$114\% SEDI@200mm) \\
P95  & 0.95  & $19\times$            & Moderate \\
P99  & 0.99  & $99\times$            & Modest ($+$3.6\% channel SEDI) \\
P999 & 0.999 & $999\times$           & Negligible ($\pm$2\% SEDI) \\
\bottomrule
\end{tabular}
\end{center}

The pattern is monotonic: the higher the asymmetric amplification from the loss function
itself, the less the head benefits from additional extreme training examples.

\paragraph{P50 is blind to extremes without augmentation.}
With symmetric loss ($1\times$), a 200\,mm event produces the same gradient magnitude
as a 2\,mm event.  The P50 head learns ``when ERA5 shows a storm, predict
${\sim}$5--10\,mm''---accurate for typical days, useless at 200\,mm.
Adding 83 synthetic extreme days provides the P50 head its only meaningful exposure
to extreme-event spatial patterns, explaining the 0.405\,$\to$\,0.866 SEDI jump at
200\,mm.

\paragraph{P999 is already saturated from real data.}
With $999\times$ asymmetric loss, each extreme pixel-day contributes a
$999\times$-amplified gradient.  Adding synthetic days increases the effective
gradient budget by only ${\sim}$10\%.  Moreover, the \cvae{}'s spatial
correlation is 0.885, not 1.0: every spatial misplacement is amplified $999\times$,
offsetting the marginal benefit with amplified noise.  The net result is near-zero
augmentation effect (SEDI at 200\,mm: 0.932\,$\to$\,0.922).

\paragraph{Why augmentation helps Q-SRDRN's P50 but not SRDRN.}
Both \srdrn{} and P50 predict a median-like quantity, yet augmentation
barely helped \srdrn{} (SEDI $+$2.3\%, KL \emph{worsened}) while dramatically
improving P50 (SEDI@200mm $+$114\%, KL $-$50\%).  Three mechanisms explain the
difference:

\begin{enumerate}[noitemsep]
\item \textbf{Intensity weighting amplifies label conflicts in SRDRN.}
  A synthetic 170\,mm pixel receives $20\times$ the gradient of a real dry pixel
  in \srdrn{}, far exceeding its 0.67\% data fraction.  P50's unweighted pinball loss
  makes real and synthetic pixels contribute equal gradient, so 83 synthetic days
  shift the median by $<$0.6\%.

\item \textbf{Gradient saturation caps SRDRN's extreme-event signal.}
  Events above 23.4\,mm produce identical gradients in \srdrn{}.
  Pinball loss has no saturation: the residual $|y - \hat{q}|$ scales linearly
  with event severity.

\item \textbf{Single output vs.\ separate heads.}
  \srdrn{}'s single output must simultaneously serve all purposes.
  \qsrdrn{}'s separate heads let augmentation improve P50's extreme-day
  predictions without affecting P999's filters.
\end{enumerate}

\paragraph{Scope of the ``P999 immune to augmentation'' claim.}
The immunity statement holds only in the regime tested here: 83 \cvae{} samples
($\sim$0.67\% of training), $\tau$\,=\,0.999, FL/CA precipitation,
\cvae{} sampling correlation 0.83--0.89.  At higher augmentation ratios or
lower-quality \cvae{}s the $999\times$ amplification becomes a liability rather
than an asymptote (Run 11 in Appendix~\ref{app:augmentation}: 1.5\% ratio
regressed all metrics).  We do not claim immunity outside this operating point.

\section{Hyperparameters}
\label{app:hyperparams}

\paragraph{Q-SRDRN backbone.}
16 residual blocks (each: two Conv2D $3\times3$ kernels with PReLU and
SpatialDropout2D(0.1)), 256 filters, asymmetric pixel-shuffle upsampling
($6\times$ lat, $4\times$ lon for FL; $6\times6$ for CA), four separate
Conv2D(1, kernel\,$=9$) output heads concatenated and passed through
IncrementBound.  Backbone weights initialised via He normal.

\paragraph{Training.}
Adam, lr $=10^{-5}$, no scheduler.  160 epochs, batch size 64 (FL).
Inputs (the 15 ERA5 surface variables enumerated in \S\ref{subsec:setup})
z-normalised per channel; targets log1p-transformed and z-normalised, inverted
at evaluation. The loss is the masked pinball objective of Eq.~\ref{eq:loss}
with $\tau \in \{0.50, 0.95, 0.99, 0.999\}$ and event weight $\alpha = 5.0$ on
the P95/P99/P999 heads ($z_\text{thresh} = 0.5$ in log1p z-score units); the
P50 head is unweighted.

\paragraph{cVAE.}
PyTorch, latent $z \in \mathbb{R}^{256}$, $\beta_\text{KL} = 0.008$,
encoder/decoder share the same 5-block backbone with GroupNorm and no
skip connections.  Adam, lr $=2\times10^{-4}$, 300 epochs, batch size 16.
FL augmentation set is 83 samples drawn from posterior $z$,
filtered by min-intensity-ratio $\ge$1.03 over the original PRISM target.

\paragraph{Postprocessing.}
Per-channel physical caps (P50/P95: 600\,mm, P99/P999: 1{,}200\,mm)
applied after IncrementBound; no sort needed (IncrementBound is monotone
by construction).  Log1p inverse and target denormalization applied last.

\section{UQ-Baseline Comparison: Detailed Results}
\label{app:uq_baselines}

This appendix contains the full results table and per-baseline diagnostics
underlying the summary in \S\ref{subsec:uq_baselines}.  All four comparators
fix \qsrdrn{}'s backbone, optimiser, training data, and 83-sample \cvae{}
augmentation, and vary only the predictive head and the mechanism that
produces tail quantiles.

\begin{table}[!htbp]
\centering
\caption{\textbf{Florida and California: \qsrdrn{} alongside complementary uncertainty-quantification
designs.}
Backbone, optimiser, training data, and 83-sample \cvae{} augmentation are held fixed
across the four comparators (identical augmentation set used by the headline
\qsrdrn{}+\cvae{} runs of Table~\ref{tab:aug}); only the head
architecture and the mechanism that produces tail quantiles vary.  Each comparator
answers a different question (does pinball replace MC sampling? does sharing all
weights across $\tau$ work? does ensembling deterministic predictors reproduce a
quantile spread?), so this table is not a head-to-head ranking---no row-best
emphasis is applied.  P999~calib.\ target is 0.10\%; values in the
20--30\% range indicate the head's predictive variance is approximately constant
across $\tau$, i.e.\ the model cannot distinguish median from extreme.}
\label{tab:uq_baselines}
\vspace{4pt}
\footnotesize
\renewcommand{\arraystretch}{1.05}
\begin{tabular}{lccccc}
\toprule
\textbf{Metric (direction)} & \textbf{\qsrdrn{}+\cvae{}} & \textbf{\textsc{multi}} & \textbf{\textsc{single}} & \textbf{\textsc{mc}} & \textbf{\textsc{deep}} \\
                            &                            & \textbf{\_seed}         & \textbf{\_head\_qr}      & \textbf{\_dropout} & \textbf{\_ensemble} \\
\midrule
\multicolumn{6}{l}{\textit{Florida}} \\
\addlinespace
RMSE ($\downarrow$, mm/day)         & 8.665        & 8.640{\scriptsize$\,\pm\,0.042$} & 8.798 & 16.726 & 8.618$^{\S}$ \\
Pearson $r$ ($\uparrow$)            & 0.629        & 0.630{\scriptsize$\,\pm\,0.005$} & 0.618 & 0.332  & 0.649$^{\S}$ \\
KL divergence ($\downarrow$)        & 0.023        & 0.022{\scriptsize$\,\pm\,0.006$} & 0.032 & 0.003$^{\dagger}$ & 0.041 \\
\addlinespace
P999 SEDI @ 200\,mm ($\uparrow$)               & 0.932 & 0.928{\scriptsize$\,\pm\,0.004$} & 0.939 & 0.888 & 0.885 \\
P999 SEDI @ 300\,mm ($\uparrow$)               & 0.735 & 0.726{\scriptsize$\,\pm\,0.035$} & 0.774$^{\ddagger}$ & 0.631 & 0.603 \\
P999 POD  @ 200\,mm ($\uparrow$, $n{=}2{,}111$) & 0.738 & 0.758{\scriptsize$\,\pm\,0.016$} & 0.795 & 0.676 & 0.564 \\
P999 hits @ 300\,mm ($\uparrow$, $n{=}75$)      & 23    & 22{\scriptsize$\,\pm\,4$}        & 29    & 19    & 8 \\
\addlinespace
P95  empirical exceed.\ (target 5\%)    & 4.93\% & 4.32\% & 4.40\% & 26.7\% & 25.6\% \\
P99  empirical exceed.\ (target 1\%)    & 1.76\% & 1.71\% & 1.62\% & 25.7\% & 25.1\% \\
P999 empirical exceed.\ (target 0.10\%) & 0.43\% & 0.47\% & 0.41\% & 25.5\% & 25.0\% \\
\midrule
\multicolumn{6}{l}{\textit{California}} \\
\addlinespace
RMSE ($\downarrow$, mm/day)         & 2.881  & ---$^{\|}$ & 2.856  & 2.994  & 2.731 \\
Pearson $r$ ($\uparrow$)            & 0.848  & ---$^{\|}$ & 0.849  & 0.837  & 0.864 \\
KL divergence ($\downarrow$)        & 0.0008 & ---$^{\|}$ & 0.0029 & 0.0012 & 0.0024 \\
\addlinespace
P999 SEDI @ 200\,mm ($\uparrow$)             & 0.993 & ---$^{\|}$ & 0.999 & 0.946 & 0.814 \\
P999 SEDI @ 300\,mm ($\uparrow$)             & 1.000 & ---$^{\|}$ & 1.000 & 0.978 & 0.787 \\
P999 POD  @ 200\,mm ($\uparrow$, $n{=}537$)  & 0.968 & ---$^{\|}$ & 0.994 & 0.743 & 0.315 \\
P999 hits @ 300\,mm ($\uparrow$, $n{=}15$)   & 15    & ---$^{\|}$ & 15    & 13    & 3 \\
\addlinespace
P95  empirical exceed.\ (target 5\%)    & 2.02\% & ---$^{\|}$ & 2.24\% & 15.0\% & 13.6\% \\
P99  empirical exceed.\ (target 1\%)    & 0.73\% & ---$^{\|}$ & 0.74\% & 14.5\% & 13.3\% \\
P999 empirical exceed.\ (target 0.10\%) & 0.16\% & ---$^{\|}$ & 0.16\% & 14.4\% & 13.3\% \\
\bottomrule
\end{tabular}
\par\vspace{4pt}
{\footnotesize\raggedright
$^{\dagger}$ \textsc{mc\_dropout}'s low KL reflects an aggregated wet-day distribution
match; the per-quantile structure (rows below) is broken.\;
$^{\ddagger}$ At 300\,mm only $n{=}75$ observed events exist; this $+0.04$ inversion
is within deep-tail sampling noise.\;
$^{\S}$ \textsc{deep\_ensemble}'s 0.5\% RMSE / 1.9\,p.p.\ Pearson edge is the
expected variance-reduction artefact of averaging 5 deterministic predictors; its
upper-quantile rows below show the cost.\;
\textsc{multi\_seed} entries report mean\,$\pm\,1\sigma$ across seeds
$\{11,23,47,71,97\}$; per-seed empirical-exceedance means are reported without
$\sigma$ (per-seed spread $<\,0.4$\,p.p.\ across all four quantiles).
The single-seed \qsrdrn{}+\cvae{} reference column reports POD and hit counts
recomputed directly from its verified test predictions; bulk metrics and the
P999-head SEDI/POD/hit rows are reproduced by the \textsc{multi\_seed} mean
within $\le$1\,$\sigma$.
$^{\|}$ California \textsc{multi\_seed} data was not ready at the time of
submission and will be populated in the camera-ready version.
All other comparators here are trained on the same 83-sample \cvae{} augmentation
set used by the reference \qsrdrn{}+\cvae{} column, so the comparison isolates
head architecture and quantile mechanism only.
\par}
\end{table}

\paragraph{MC-Dropout: predictive variance is approximately quantile-invariant.}
The \textsc{mc\_dropout} columns in Table~\ref{tab:uq_baselines} reveal a
single mechanism behind two failure modes.  Empirical exceedance rates at
P95, P99, and P999 are 26.7\%, 25.7\%, and 25.5\%---essentially identical,
and 5--250$\times$ above their respective targets.  Dropout's predictive
variance is a near-constant function of the input rather than a function of
$\tau$: every quantile collapses onto roughly the same wet-day band.  The
same effect drives the bulk-RMSE blow-up: noisy $K$-pass averaging
hallucinates extremes, nearly doubling RMSE from 8.665 to 16.726.  This
is a strong domain-specific demonstration
of the MC-Dropout failure mode flagged by \citet{gal2016dropout} for tail
prediction: \emph{a 250$\times$ overshoot at the 99.9th-percentile target on
a real precipitation downscaling task}.  The same quantile-invariance failure
mode reproduces on California (P95/P99/P999 empirical exceedance
$=$ $15.0\%$, $14.5\%$, $14.4\%$ --- 3$\times$, 14$\times$, 144$\times$ above
their respective targets), with a milder bulk-RMSE penalty ($+4\%$ vs.\
\qsrdrn{}+\cvae{}) than on Florida because California's lighter convective
tail gives the noisy empirical median less room to hallucinate extremes.
Explicit pinball regression is therefore not optional for extreme-quantile
prediction.

\paragraph{Single-head QR-CNN: bulk-competitive, but the median cannot reach extremes.}
\textsc{single\_head\_qr} is the strongest comparator on bulk metrics: RMSE
within $1.5\%$ of \qsrdrn{}, Pearson within $0.011$, per-channel calibration
within $0.6$\,p.p.\ of \qsrdrn{}+\cvae{}, and deep-tail P999 detection at
$300$\,mm slightly above the \qsrdrn{} reference in both SEDI ($0.774$ vs.\
$0.735$) and raw hits ($29$ vs.\ $23$ of $n{=}75$). At $n{=}75$ this
$+0.04$ SEDI inversion sits within deep-tail sampling spread; the post-hoc
\texttt{np.sort} is sufficient to enforce non-crossing once the network has
learned roughly correct quantiles, and the P999 head's own gradient signal
does most of the work for ultra-extreme detection regardless of whether
channels are shared. The cost of shared-head \texttt{tf.sort} shows up
indirectly in the predictive distribution's \emph{shape}: KL rises by
$41\%$ ($0.023 \to 0.032$) and the sharpness table
(Appendix~\ref{app:ablation}) shows P99$-$P50 medians widening to
$38$--$138$\,mm for \textsc{single\_head\_qr} vs.\ $8$--$22$\,mm for
\qsrdrn{}+\cvae{}, the empirical signature of the gradient-permutation
pathology of \S\ref{subsec:ablation}---per-pixel \texttt{tf.sort} re-routes
gradients across channels, so no filter specializes. On California, where
the conditional median is far more predictable owing to the
atmospheric-river regime (\S\ref{subsec:architecture}),
\textsc{single\_head\_qr} and \qsrdrn{}+\cvae{} are essentially
indistinguishable on bulk metrics (RMSE $2.856$ vs.\ $2.881$; Pearson
$0.849$ vs.\ $0.848$) and on P999 detection (P999 SEDI@200\,mm $0.999$
vs.\ $0.993$; both reach 15/15 hits at 300\,mm).

\paragraph{Deep ensemble: same quantile-invariant variance, different mechanism.}
\textsc{deep\_ensemble} clusters tightly across seeds (per-member RMSE std
$=0.09$\,mm), and the 5-member mean attains the lowest bulk RMSE in
Table~\ref{tab:uq_baselines} (8.618 vs.\ 8.665), the standard
variance-reduction artefact of averaging multiple trained models.  But
upper-quantile calibration collapses in exactly the same shape as
\textsc{mc\_dropout}: P95/P99/P999 empirical exceedance is 25.6/25.1/25.0\%,
again 5--250$\times$ above target.  The mechanism differs---$M{=}5$
deterministic predictions cannot span the 0.999 quantile rank by
construction---but the symptom is identical, and the pinball CRPS at the
$[200,\infty)$ band (68.8) is the worst of any comparator (vs.\
\textsc{mc\_dropout}'s 52.8 and \textsc{single\_head\_qr}'s 54.2).  P999
SEDI at 300\,mm degrades to 0.603, a moderate loss relative to
\qsrdrn{}+\cvae{}.  Together, \textsc{mc\_dropout}
and \textsc{deep\_ensemble} demonstrate that \emph{empirical-quantile
aggregation of point predictors cannot calibrate the deep tail}, regardless
of whether the source of spread is dropout noise ($K{=}30$) or seed variance
($M{=}5$).  This is the central architectural argument of the section:
explicit per-quantile pinball regression directly optimises the target loss
at each $\tau$ and is not substitutable by post-hoc empirical-quantile
wrappers.

\paragraph{Multi-seed: Q\_SRDRN+cVAE is a typical seed, not a lucky one.}
\textsc{multi\_seed} fixes the \qsrdrn{} architecture and 83-sample
augmentation and varies only the random seed across $\{11,23,47,71,97\}$.
All five seeds land within a tight band: RMSE $8.640\,\pm\,0.042$ (0.5\%
spread), Pearson $0.630\,\pm\,0.005$, KL $0.022\,\pm\,0.006$,
P999~SEDI@200\,mm $0.928\,\pm\,0.004$, and P999~SEDI@300\,mm
$0.726\,\pm\,0.035$.  The 5-seed mean is statistically indistinguishable
from the Q\_SRDRN+cVAE reference column on every metric the architecture is designed
to optimise (RMSE within 1\,$\sigma$, Pearson within 0.2\,$\sigma$, KL
within 0.2\,$\sigma$, P999~SEDI@300\,mm within 0.3\,$\sigma$), and matches
or slightly exceeds Q\_SRDRN+cVAE on KL ($0.022$ vs.\ $0.023$).  Critically,
per-quantile calibration is preserved across every seed---empirical
exceedance is $4.32$/$1.71$/$0.47$\% at P95/P99/P999 (target
$5/1/0.1$\%), in the same regime as Q\_SRDRN+cVAE
and \textsc{single\_head\_qr}, and nowhere near the $\approx\,25$\%
quantile-invariant collapse of \textsc{mc\_dropout} and
\textsc{deep\_ensemble}.  This contrast is mechanism-revealing: averaging
\emph{already-calibrated pinball heads} across $M{=}5$ seeds preserves
quantile structure, whereas extracting empirical quantiles \emph{from the
spread of point predictors} (whether dropout passes or ensemble members)
destroys it.  The headline architectural argument therefore holds across
the spread of \qsrdrn{} itself.

\paragraph{California.}
Three of the four comparators are populated in Table~\ref{tab:uq_baselines};
the \textsc{multi\_seed} data was not ready at the time of submission and
will be populated in the camera-ready version.  Two findings are
already visible.  First, the \textsc{mc\_dropout} quantile-invariance failure
mode reproduces on California --- P95/P99/P999 empirical exceedance $=$
$15.0\%$, $14.5\%$, $14.4\%$ ($3{\times}$, $14{\times}$, $144{\times}$ above
target) --- with a much smaller bulk-RMSE penalty ($+4\%$ vs.\ \qsrdrn{}+\cvae{})
than on Florida, because California's lighter convective tail gives the
noisy empirical median less room to hallucinate extremes; the calibration
mechanism is the same, the bulk amplification is regime-driven.  Second,
\textsc{deep\_ensemble} reproduces the same empirical-quantile-aggregation
ceiling as on Florida --- P95/P99/P999 saturate at $13.6\%$, $13.3\%$,
$13.3\%$ ($133{\times}$ above the P999 target) --- confirming that
$M{=}5$ deterministic predictors cannot span the predictive tail in either
regime, even when their per-member bulk fit is competitive (RMSE $2.731$ vs.\
\qsrdrn{}+\cvae{}'s $2.881$).

\section{Probabilistic Calibration Diagnostics: Detailed Results}
\label{app:calibration}

This appendix contains the full reliability, CRPS, and sharpness diagrams
underlying the calibration summary in \S\ref{subsec:calibration}.

\begin{figure}[t]
  \centering
  \includegraphics[width=\textwidth]{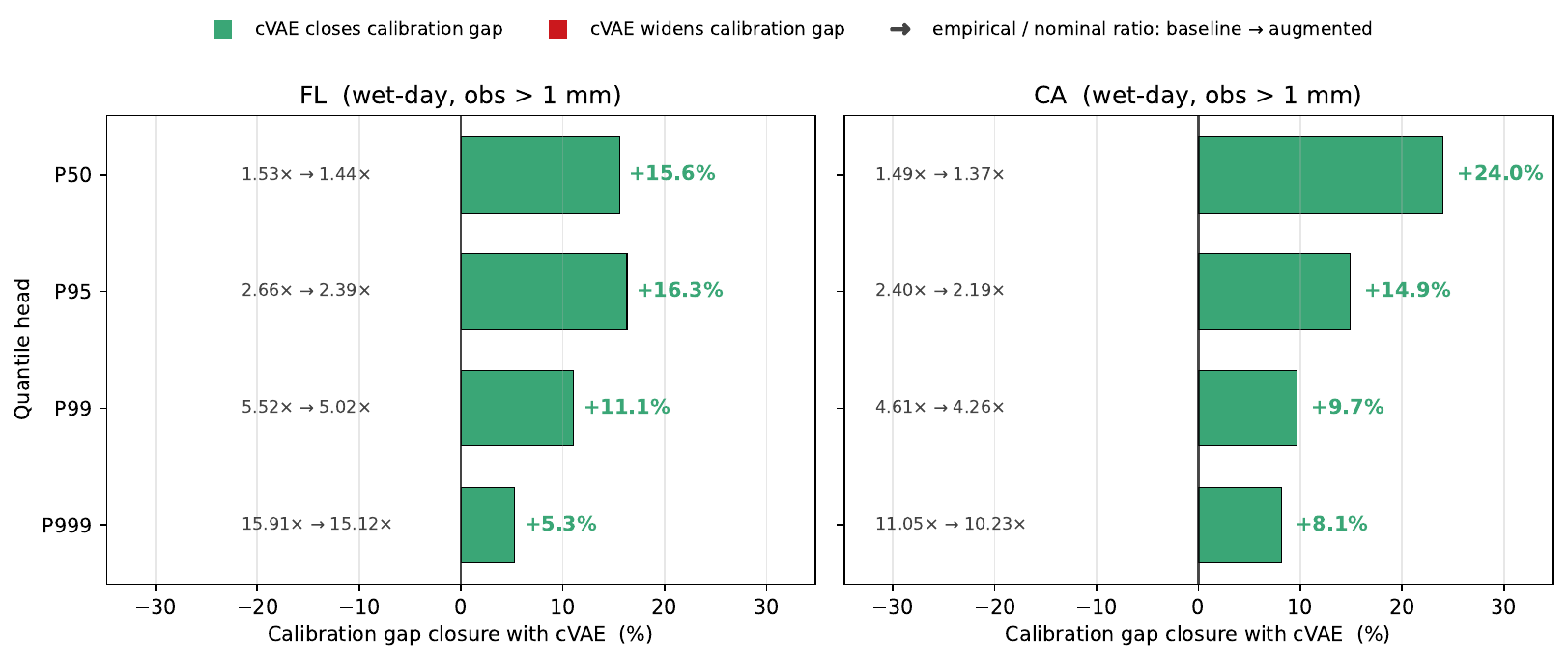}
  \caption{\textbf{Calibration gap closure with \cvae{} augmentation
  (wet days, $y > 1$\,mm).}
  Bar length is the relative reduction in calibration gap from \cvae{}
  augmentation: $\text{gap closure} = 1 - (r_\text{aug} - 1) / (r_\text{base} - 1)$,
  where $r = \Pr(y > \hat q_\tau)/(1-\tau)$ is the empirical-to-nominal
  exceedance ratio (a value of $1$ is perfect calibration).  Positive (green)
  bars mean augmentation moves the head toward perfect calibration; negative
  bars would mean it moves away.  Annotations to the left of each bar report
  the absolute baseline and augmented ratios, e.g.\ ``$1.53\times \to 1.44\times$''
  for FL P50 means the model over-predicts wet-day exceedance by $1.53\times$
  without augmentation and $1.44\times$ with it.  All 8/8 head$\times$domain
  combinations show positive gap closure ($5$--$24\%$, mean $13\%$).  Under-prediction
  widens monotonically with $\tau$ ($\sim$1.5$\times$ at P50, $\sim$16$\times$ at
  P999), motivating IncrementBound and per-quantile event weighting.  We plot
  the gap closure rather than the raw ratios on a log axis because the ratios
  span $1.5\times$ to $16\times$ and visually crush percentage-level changes
  from augmentation into invisibility.}
  \label{fig:reliability}
\end{figure}

\paragraph{Reliability.}
Fig.~\ref{fig:reliability} reports calibration gap closure from \cvae{}
augmentation for each of the four quantile heads, restricted to wet days
($y > 1$\,mm) to remove the zero-inflation that makes all-sky exceedance
uninterpretable for $\tau = 0.5$.  Defining the calibration ratio
$r = \Pr(y > \hat q_\tau)\,/\,(1-\tau)$, $r = 1$ is perfect calibration and
$r > 1$ means the model under-predicts the quantile.  All eight head/domain
combinations have $r > 1$ at baseline, and $r$ widens monotonically with
$\tau$ (FL: $1.53\times$ at P50, $5.52\times$ at P99, $15.91\times$ at P999;
CA: $1.49\times$ at P50, $11.05\times$ at P999), the residual under-prediction
that the architectural design of \S\ref{subsec:qsrdrn} aims to compress.
The bar lengths in Fig.~\ref{fig:reliability} report the relative reduction
in calibration gap, $1 - (r_\text{aug} - 1)/(r_\text{base} - 1)$.  All 8/8
head$\times$domain combinations show positive gap closure with magnitudes
$5$--$24\%$ (mean $13\%$; largest on CA P50 at $+24\%$ and FL P95 at $+16\%$).
Crucially, this is gap closure on top of \emph{not} breaking wet-day
calibration on either domain---including California, where augmentation hurts
P50 extreme \emph{detection}.  The all-sky P50 exceedance of ${\sim}$32\% on
FL and ${\sim}$18\% on CA is \emph{not} miscalibration: ${\sim}$66\% (FL)
and ${\sim}$78\% (CA) of pixel-days are dry, so the true conditional median
is exactly 0\,mm and the all-sky exceedance reduces to the probability of
precipitation, not 50\%.  The $[\text{P50}, \text{P99}]$ interval coverage
confirms joint calibration: FL 46.9\% (target 49\%), CA 48.0\%, both
near-ideal.

\begin{figure}[t]
  \centering
  \includegraphics[width=\textwidth]{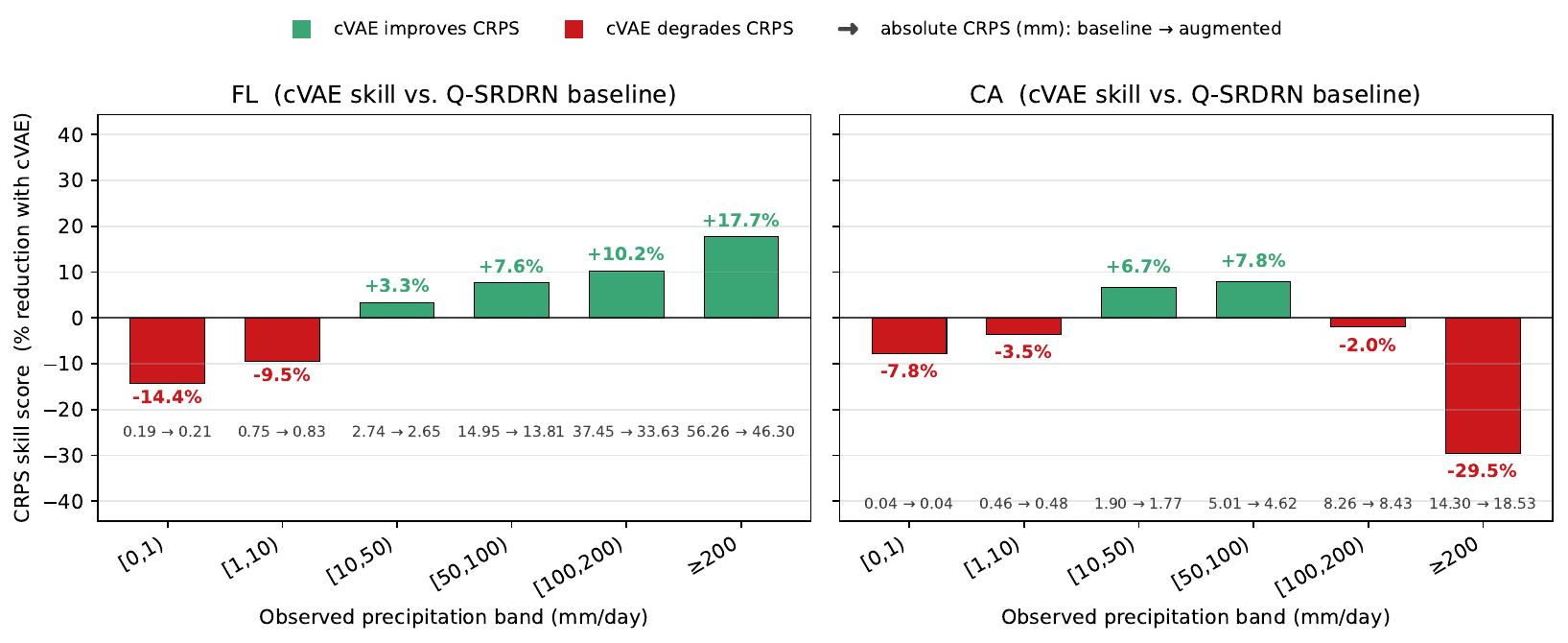}
  \caption{\textbf{CRPS skill score by observed intensity band.}
  Bars show $s = (\mathrm{CRPS}_\text{base} - \mathrm{CRPS}_\text{aug})/\mathrm{CRPS}_\text{base}$
  per band: positive (green) bars mean \cvae{} augmentation improves the pinball-CRPS proxy;
  negative (red) bars mean it hurts.  Underlying CRPS values (mm/day) are
  printed beneath each bar so the absolute magnitude is recoverable.  On Florida,
  augmentation produces positive skill in every band $\ge$10\,mm/day, peaking at
  $+17.7\%$ in the heaviest band; on California the skill is positive in the mid
  bands [10,100) but flips to $-29.5\%$ in the heaviest band---a regime-dependent
  signature consistent with Sec.~\ref{subsec:decomposition}.  We use the skill
  score rather than absolute log bars because the proxy spans almost three orders
  of magnitude across bands, which makes the (small but consequential) percentage
  changes invisible on a log axis.}
  \label{fig:crps}
\end{figure}

\begin{figure}[t]
  \centering
  \includegraphics[width=\textwidth]{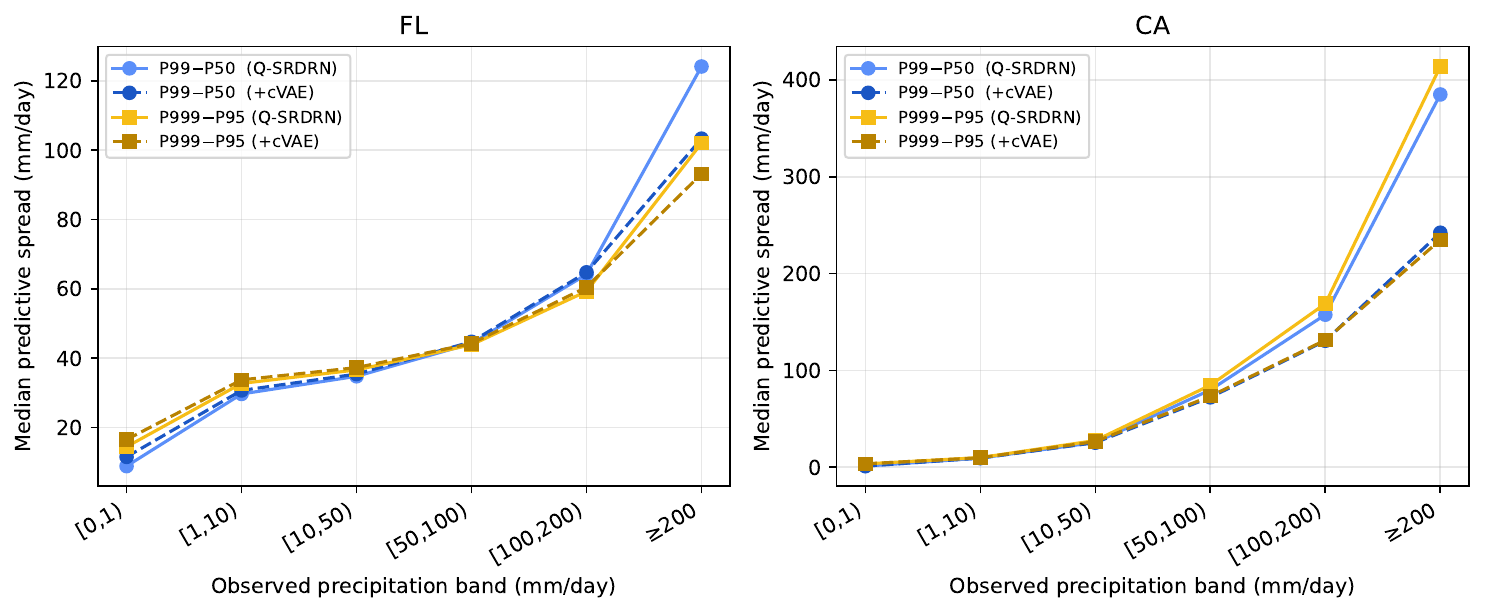}
  \caption{\textbf{Sharpness conditioned on observed intensity.}
  Median predictive spread P99\,$-$\,P50 (circles) and P999\,$-$\,P95 (squares),
  binned by observed precipitation.  Spread grows monotonically with intensity
  on both domains, confirming that the upper quantile heads stretch where the
  data demands.  cVAE augmentation narrows the predictive spread (raises
  sharpness) on the extreme bands of \emph{both} domains, and the effect is
  substantially larger on CA than on FL.  Sharpness therefore does not mirror
  the CRPS-skill split of Fig.~\ref{fig:crps}: the same CA sharpening coincides
  with negative CRPS skill in the heaviest band, so the added concentration is
  uncalibrated rather than informative (consistent with the P50@200\,mm SEDI
  loss in Table~\ref{tab:contributions}).}
  \label{fig:sharpness}
\end{figure}

\paragraph{CRPS skill and sharpness.}
The decisive evidence for the differential augmentation effect is in
Fig.~\ref{fig:crps}, which reports a CRPS skill score per observed-intensity
band,
$s = (\mathrm{CRPS}_\text{base} - \mathrm{CRPS}_\text{aug})/\mathrm{CRPS}_\text{base}$,
where the underlying score is the pinball-CRPS proxy
$\mathrm{CRPS}_{\text{proxy}} = \tfrac{1}{4}\sum_\tau \rho_\tau(\hat q_\tau, y)$
which on an even quantile grid is proportional to CRPS~\citep{gneiting2007strictly}.
Plotting the skill score (rather than the raw proxy on a log axis, which
compresses all relative changes) makes the regime split of
Sec.~\ref{subsec:decomposition} quantitative.  On Florida, augmentation
improves CRPS-proxy in every band $\ge$10\,mm, with the largest gain in the
heaviest band ($56.3 \to 46.3$\,mm/day at obs $\ge$200\,mm/day, $+17.7\%$
skill).  On California the score is positive on the mid-intensity bands
[10,100) (up to $+7.8\%$) but \emph{negative} in the heaviest band
($14.3 \to 18.5$\,mm/day, $-29.5\%$ skill)---direct probabilistic evidence
for the same regime-dependent role of \cvae{} that the threshold-detection
table already showed.  Fig.~\ref{fig:sharpness} reports the underlying
sharpness directly.  Median predictive spread (P99\,$-$\,P50 and
P999\,$-$\,P95) grows monotonically with observed intensity on both domains,
confirming the upper quantile heads stretch where the data demands.  cVAE
augmentation narrows the spread (raises sharpness) on the extreme bands of
\emph{both} domains---substantially more on CA than on FL---so sharpness
alone does not mirror the CRPS-skill split.  The CA contrast is the
informative one: the same augmentation that sharpens the predictive
distribution there also costs CRPS skill in the heaviest band, so the
additional concentration is uncalibrated rather than informative, consistent
with the P50@200\,mm SEDI loss in Table~\ref{tab:contributions}.

\section{Limitations: Detailed Discussion}
\label{app:limitations_extra}

\paragraph{Generator quality: \cvae{} versus conditional diffusion.}
The augmentation generator in this paper is a \cvae{} with sampling spatial
correlations of $0.885$ on Florida and $0.707$ (vanilla) / $0.835$ (with
AdaIN) on California---the latter still well below $1.0$. cVAEs are known
to produce smoother, lower-variance samples than state-of-the-art
conditional diffusion models because of their Gaussian-posterior
parameterisation; spatial misplacements in the synthetic fields are then
amplified by the $999{\times}$ pinball gradient at the deep tail
(Appendix~\ref{app:aug_mechanism}), which is why \cvae{} augmentation is
neutral on the P999 head in the regime tested here. A higher-fidelity
generator---e.g., a physics-constrained conditional diffusion model---could
plausibly produce samples crisp enough to lift P99 and P999 detection
beyond what real data alone supports. The natural design is a \emph{hybrid}:
diffusion is used only offline to expand the training set, while \qsrdrn{}
retains its single-pass inference. This separates the (expensive)
sample-generation step from the (cheap) downscaling step, preserving the
computational profile that distinguishes \qsrdrn{} from end-to-end
diffusion downscalers.

\paragraph{Residual calibration at the deep tail.}
The reliability diagnostic of \S\ref{subsec:calibration} closes 5--24\% of the
calibration gap across all eight head--domain combinations, but the P999 head
retains a $15.12{\times}$ wet-day under-prediction ratio on Florida; the
architecture compresses the residual gap without eliminating it. Two
gradient-side extensions are natural candidates for closing it further.
First, the event weight $\alpha=5$ is fixed during training; making $\alpha$
\emph{dynamic}---scaling with the running batch variance of pinball loss at
each head, or with the empirical exceedance gap accumulated over a sliding
window---would let the upper-tail heads receive proportionally more gradient
when calibration is drifting and less when it is on target. Second, a
Focal-Loss-inspired dynamic scaling on the pinball residual
(down-weighting easily-predicted residuals and up-weighting hard ones)
would concentrate optimisation pressure on the deep tail. Both modifications
operate at the gradient level only, compose with IncrementBound and the
separate-heads architecture without changing head topology, and leave the
quantile set $\mathcal{T}$ intact; we leave their empirical validation to
future work.

\paragraph{Broader applicability.}
The \qsrdrn{} framework is architecture-agnostic: IncrementBound and
per-quantile event weighting can be applied to any CNN backbone (U-Net,
diffusion models) and to any right-skewed geophysical variable (wind gusts,
streamflow, wave height). Demonstrating transfer to those settings is the
clearest route to establishing whether the heavy-tail failure mode we
diagnose is general or precipitation-specific.

\section{Per-Component Contributions Table}
\label{app:contributions}

This appendix decomposes the headline numbers of \S\ref{sec:experiments}
into the additive contribution of each component: the architecture switch
\srdrn{}$\to$\qsrdrn{} (column \textit{Arch.}) and the further
\cvae{} augmentation on top of \qsrdrn{} (column \textit{cVAE}). All
values in Table~\ref{tab:contributions} are single-seed test-set deltas
signed in the improvement direction. The architecture switch dominates
the extreme-tail metrics (P999 SEDI@300\,mm: $+0.71$ Florida, $+1.00$
California, $+0.89$ Texas substate), while \cvae{} augmentation provides
the largest median lift (P50 SEDI@200\,mm: $+0.46$ Florida) and is
region-specific (\S\ref{subsec:augmentation}).

\begin{table}[H]
\centering
\caption{\textbf{Per-component contributions, by metric and domain.}
\textit{Arch.} compares \qsrdrn{} (no aug) to \srdrn{};
\textit{cVAE} compares \qsrdrn{}+\cvae{} to \qsrdrn{}.
Positive values indicate improvement in the listed direction.
Numbers are single-seed test-set deltas and should be read as point estimates,
not significance-tested effects.  Texas substate \cvae{}-column entries are
not reported; the FL-tuned \cvae{} configuration was not run there
(\S\ref{subsec:augmentation}, Table~\ref{tab:aug}).}
\label{tab:contributions}
\vspace{4pt}
\footnotesize
\renewcommand{\arraystretch}{1.05}
\begin{tabular}{lcc|cc|cc}
\toprule
& \multicolumn{2}{c|}{\textbf{Florida}} & \multicolumn{2}{c|}{\textbf{California}} & \multicolumn{2}{c}{\textbf{Texas substate}} \\
\textbf{Metric (direction)}             & \textbf{Arch.} & \textbf{cVAE} & \textbf{Arch.} & \textbf{cVAE} & \textbf{Arch.} & \textbf{cVAE} \\
\midrule
RMSE  ($\downarrow$, mm/day)            & $-$0.358 & $-$0.139 & $-$0.047 & $-$0.016 & $-$1.128 & --- \\
Pearson $r$ ($\uparrow$)                & $+$0.032 & $-$0.002 & $+$0.002 & $+$0.003 & $+$0.074 & --- \\
KL divergence ($\downarrow$)            & $-$0.078 & $-$0.023 & $-$0.0028 & $-$0.0024 & $-$0.162 & --- \\
P50 SEDI @ 200\,mm ($\uparrow$)         & $-$0.159 & $+$0.461 & $+$0.128 & $-$0.094 & $+$0.023 & --- \\
P999 SEDI @ 300\,mm ($\uparrow$)        & $+$0.710 & $+$0.025 & $+$1.000 & $\pm$0.000 & $+$0.891 & --- \\
\bottomrule
\end{tabular}
\end{table}

\section{P50 SEDI Envelope Figure}
\label{app:p50_sedi_fig}

\begin{figure}[H]
  \centering
  \begin{subfigure}[t]{0.48\textwidth}
    \includegraphics[width=\textwidth]{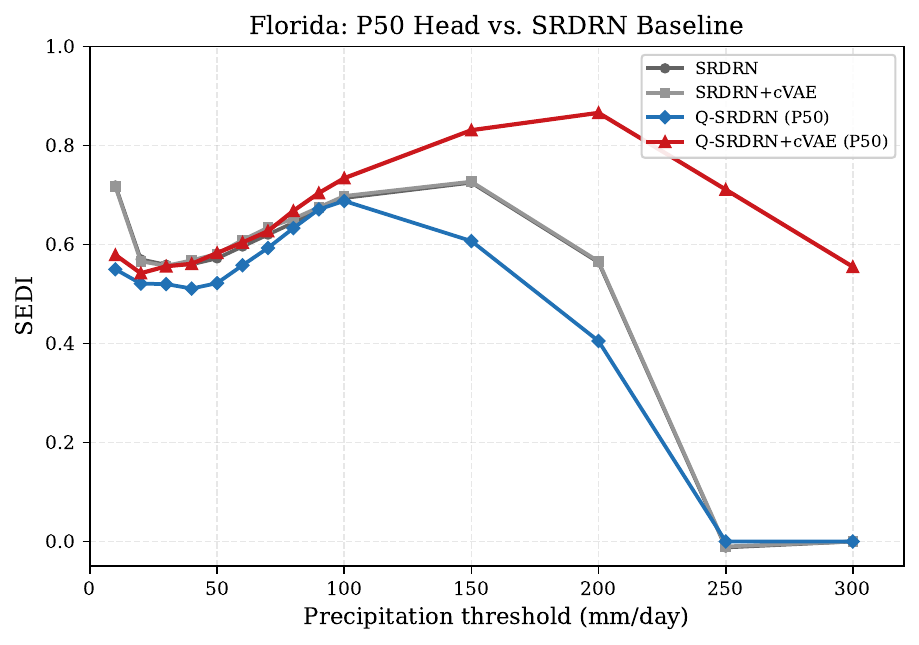}
    \caption{Florida}
  \end{subfigure}
  \hfill
  \begin{subfigure}[t]{0.48\textwidth}
    \includegraphics[width=\textwidth]{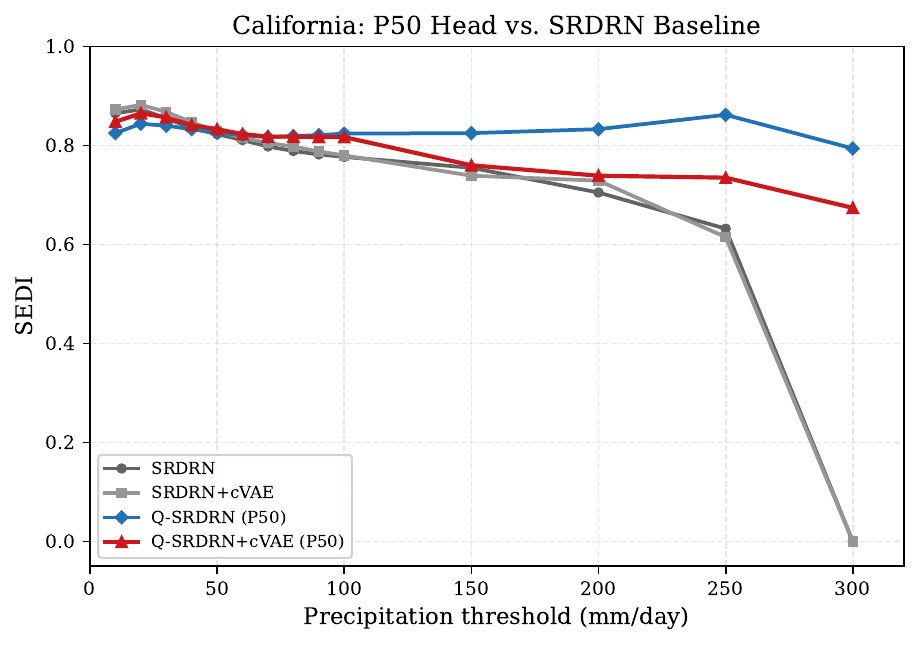}
    \caption{California}
  \end{subfigure}
  \caption{\textbf{P50 SEDI across precipitation thresholds.}
  \srdrn{} (single output) vs.\ \qsrdrn{} P50.  On CA, P50 alone surpasses
  \srdrn{} at $\ge$100\,mm; on FL the un-augmented P50 trails \srdrn{} at
  extreme thresholds---a gap closed by augmentation
  (\S\ref{subsec:augmentation}).}
  \label{fig:p50_sedi}
\end{figure}

This appendix supplies the per-domain median-head SEDI envelope referenced
in \S\ref{subsec:augmentation}: the SEDI of the \qsrdrn{} P50 head as a
function of precipitation threshold, compared against the deterministic
\srdrn{} baseline. Figure~\ref{fig:p50_sedi} shows that on California the
un-augmented \qsrdrn{} P50 already surpasses \srdrn{} at thresholds
$\ge100$\,mm/day --- the median-quantile loss is sufficient on a smooth
synoptic regime --- whereas on Florida the un-augmented P50 trails
\srdrn{} at extreme thresholds, with the gap closed only by \cvae{}
augmentation.

\section{Matched-Head Envelope Figure}
\label{app:matched_head_fig}

\begin{figure}[H]
  \centering
  \includegraphics[width=\textwidth]{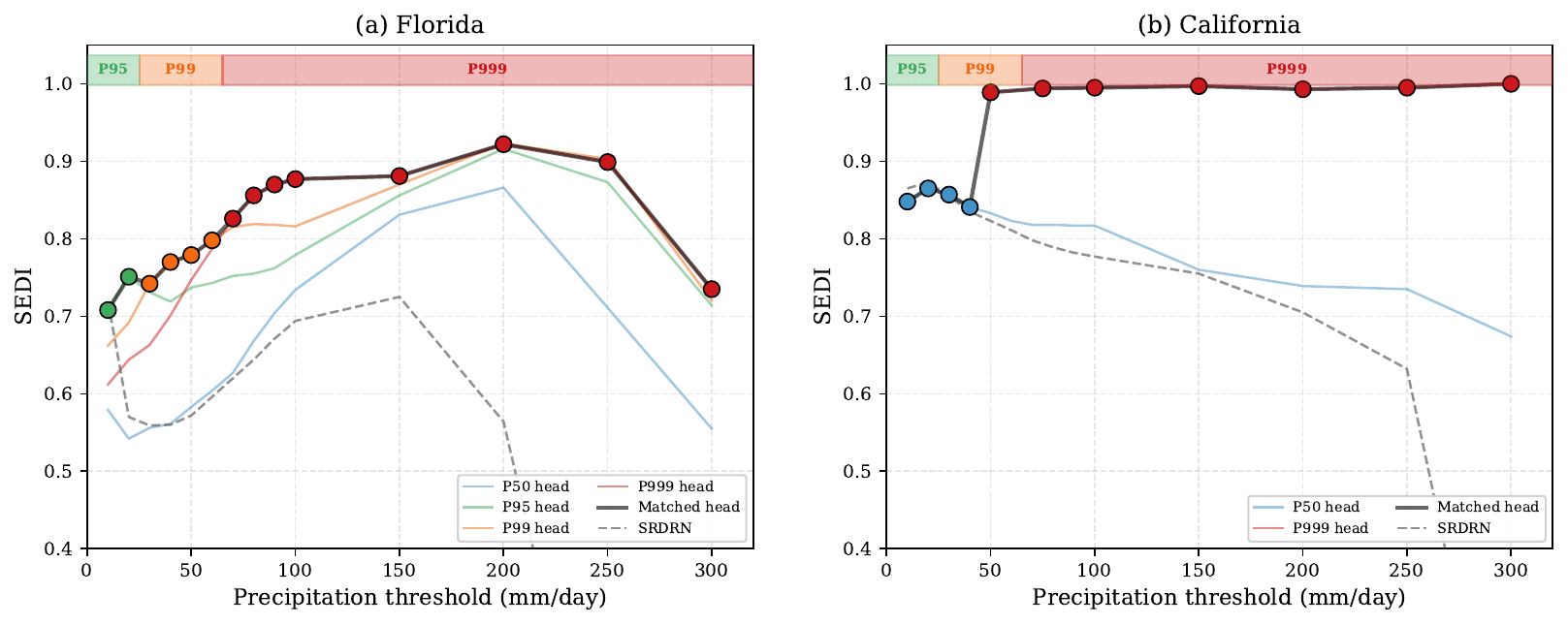}
  \caption{\textbf{Each quantile head in its natural range} (\qsrdrn{}, no augmentation).
  Head assignment follows climatological exceedance rates: P50 (blue, $\le$10\,mm),
  P95 (green, 10--20\,mm), P99 (orange, 30--60\,mm), P999 (red, $\ge$70\,mm).
  The matched-head envelope (black) exceeds the \srdrn{} baseline
  (dashed gray) at every threshold on both domains.
  On California, the P999 head achieves SEDI\,$\ge$\,0.99 across the entire
  $\ge$50\,mm range.}
  \label{fig:best_head}
\end{figure}

This appendix shows the per-head SEDI envelope of \qsrdrn{} (no
augmentation) with each quantile head plotted only over its natural
climatological range: P50 for $\le10$\,mm, P95 for $10$--$20$\,mm, P99
for $30$--$60$\,mm, and P999 for $\ge70$\,mm. Figure~\ref{fig:best_head}
demonstrates that the matched-head envelope (black) lies above the
deterministic \srdrn{} baseline (dashed grey) at every threshold on both
Florida and California, and that the California P999 head sustains
SEDI\,$\ge\,0.99$ across the entire $\ge\!50$\,mm range.

\section{Seasonal RMSE}
\label{app:seasonal}

This appendix breaks the headline RMSE numbers down by meteorological
season (DJF, MAM, JJA, SON) for each of the four configurations evaluated
in the paper. Table~\ref{tab:seasonal} shows that \qsrdrn{}+\cvae{}
attains the lowest seasonal RMSE in 7 of 8 Florida and California
season pairs, and that on the un-augmented Texas substate the
\qsrdrn{} P50 head improves over \srdrn{} in all four seasons (17--49\%
RMSE reduction), confirming that the architecture-driven gain is not a
single-season artefact.

\begin{table}[H]
\centering
\caption{\textbf{Seasonal RMSE (mm/day).}  \qsrdrn{}+\cvae{} achieves the
lowest seasonal RMSE in 7 of 8 Florida and California season pairs; on the
un-augmented Texas substate, the \qsrdrn{} P50 head beats \srdrn{} in all
four seasons (17--49\% RMSE reduction).}
\label{tab:seasonal}
\vspace{4pt}
\footnotesize
\renewcommand{\arraystretch}{1.05}
\begin{tabular}{lcccc}
\toprule
 & \textbf{\srdrn{}} & \textbf{\srdrn{}+\cvae{}} & \textbf{\qsrdrn{}} & \textbf{\qsrdrn{}+\cvae{}} \\
\midrule
\multicolumn{5}{l}{\textit{Florida}} \\
\quad DJF & 1.016 & \best{0.719} & 1.240 & 0.977 \\
\quad MAM & 1.329 & 1.180 & 1.240 & \best{1.055} \\
\quad JJA & 3.478 & 3.187 & 3.677 & \best{2.844} \\
\quad SON & 1.719 & 1.822 & 1.730 & \best{1.209} \\
\midrule
\multicolumn{5}{l}{\textit{California}} \\
\quad DJF & 0.614 & 0.712 & 0.677 & \best{0.506} \\
\quad MAM & 0.413 & 0.661 & 0.435 & \best{0.326} \\
\quad JJA & \best{0.136} & 0.139 & 0.188 & 0.172 \\
\quad SON & 0.306 & 0.326 & 0.337 & \best{0.254} \\
\midrule
\multicolumn{5}{l}{\textit{Texas substate}$^{*}$} \\
\quad DJF & 1.538 & --- & \best{1.276} & --- \\
\quad MAM & 3.574 & --- & \best{1.857} & --- \\
\quad JJA & 2.706 & --- & \best{1.997} & --- \\
\quad SON & 2.365 & --- & \best{1.215} & --- \\
\bottomrule
\end{tabular}
\par\vspace{4pt}
{\footnotesize\raggedright
$^{*}$ Texas substate is reported without \cvae{} augmentation; see
\S\ref{subsec:augmentation}, Table~\ref{tab:aug}.
\par}
\end{table}

\newpage
\input{checklist.tex}

\end{document}

%% file: checklist.tex
\section*{NeurIPS Paper Checklist}

\begin{enumerate}

\item {\bf Claims}
    \item[] Question: Do the main claims made in the abstract and introduction accurately reflect the paper's contributions and scope?
    \item[] Answer: \answerYes{}
    \item[] Justification: The abstract and introduction precisely state our three contributions --- (i) we identify a CNN-specific failure mode in sort-based quantile monotonicity and resolve it with Q-SRDRN (IncrementBound + separate Conv2D(1) heads); (ii) we validate the same architecture and hyperparameters across three climatologically distinct domains (Florida, California, and a Texas Gulf-Coast substate); and (iii) we show that multi-quantile regression is what makes cVAE augmentation useful on Florida by shielding the median from the label-averaging that poisons MAE-trained networks --- and these are exactly what the experimental sections evaluate.

\item {\bf Limitations}
    \item[] Question: Does the paper discuss the limitations of the work performed by the authors?
    \item[] Answer: \answerYes{}
    \item[] Justification: Section 5.3 (Limitations) and the augmentation discussion in Section 4.5 explicitly address (i) the ERA5 25\,km resolution wall that caps deep-tail (300\,mm/day) detection regardless of model capacity, demonstrated by a capacity-expansion ablation; (ii) cVAE augmentation is region-specific (the Florida-tuned generator architecture and input-variable set mildly harm California's P50 channel and were not run on the Texas substate); (iii) cVAE generator smoothness vs.\ conditional diffusion, and the residual P999 wet-day calibration gap ($15.12\times$ on Florida); (iv) transfer to non-precipitation right-skewed geophysical variables is unverified; and (v) headline numbers are single-seed, with FL seed variance bounded by the 5-seed \textsc{multi\_seed} comparator and CA \textsc{multi\_seed} deferred to camera-ready.

\item {\bf Theory assumptions and proofs}
    \item[] Question: For each theoretical result, does the paper provide the full set of assumptions and a complete (and correct) proof?
    \item[] Answer: \answerNA{}
    \item[] Justification: The paper presents an empirical methodology (Q-SRDRN architecture, IncrementBound layer, cVAE augmentation) with no formal theorems or proofs; all results are experimental.

    \item {\bf Experimental result reproducibility}
    \item[] Question: Does the paper fully disclose all the information needed to reproduce the main experimental results of the paper to the extent that it affects the main claims and/or conclusions of the paper (regardless of whether the code and data are provided or not)?
    \item[] Answer: \answerYes{}
    \item[] Justification: Section 5.4 (Broader Impacts and Reproducibility) and the Hyperparameters appendix specify the full SRDRN/Q-SRDRN architecture, IncrementBound layer, cVAE v5c configuration, optimiser/schedule, batch sizes, multi-seed protocol, and dataset preprocessing (ERA5 15 variables, PRISM target, log1p + per-pixel normalization, train/test split). Both ERA5 and PRISM are publicly available, so an external reader has every detail required to rebuild the pipeline from scratch.

\item {\bf Open access to data and code}
    \item[] Question: Does the paper provide open access to the data and code, with sufficient instructions to faithfully reproduce the main experimental results, as described in supplemental material?
    \item[] Answer: \answerNo{}
    \item[] Justification: Code is not released at submission time but will be made publicly available upon acceptance. The paper and appendix nonetheless describe all architectural, training, and preprocessing details (Section 5.4 Broader Impacts and Reproducibility, Hyperparameters appendix) sufficient for an independent re-implementation, and both underlying data sources --- ERA5 (Copernicus CDS) and PRISM (Oregon State University PRISM Climate Group) --- are publicly downloadable.

\item {\bf Experimental setting/details}
    \item[] Question: Does the paper specify all the training and test details (e.g., data splits, hyperparameters, how they were chosen, type of optimizer) necessary to understand the results?
    \item[] Answer: \answerYes{}
    \item[] Justification: Section 4 (Experiments) and the Hyperparameters appendix specify the train/test split, Adam optimizer with documented learning rates and schedules, batch sizes (64 for SRDRN/Q-SRDRN, 48 for cVAE), epoch counts, loss-function weights (pinball quantiles, $\alpha=5$ event weighting, KL $\beta=0.008$), normalization scheme, and how each hyperparameter was selected (architectural ablations and the failed-experiments table report the search trajectory).

\item {\bf Experiment statistical significance}
    \item[] Question: Does the paper report error bars suitably and correctly defined or other appropriate information about the statistical significance of the experiments?
    \item[] Answer: \answerYes{}
    \item[] Justification: The Florida Comparison Against Uncertainty-Quantification Baselines table (Section 4.4) reports 5-seed mean $\pm$ standard deviation across seeds $\{11, 23, 47, 71, 97\}$, capturing variability from initialization and stochastic optimization, and confirms the headline UQ findings are statistically robust. Other tables and figures show single-seed results due to compute constraints; this scope is acknowledged in Section 5.3 (Limitations).

\item {\bf Experiments compute resources}
    \item[] Question: For each experiment, does the paper provide sufficient information on the computer resources (type of compute workers, memory, time of execution) needed to reproduce the experiments?
    \item[] Answer: \answerYes{}
    \item[] Justification: The Reproducibility content within Section 5.4 (Broader Impacts and Reproducibility) reports the GPU type (single NVIDIA H100 80\,GB) and per-run wall-clock time for every model and region: SRDRN baseline $\sim$6\,h (FL and TX) / $\sim$13\,h (CA); Q-SRDRN $\sim$12\,h (FL and TX) / $\sim$22\,h (CA); \cvae{} $\sim$4\,h (FL) / $\sim$8\,h (CA). The Hyperparameters appendix lists batch size, optimizer, and other run-level details.
    
\item {\bf Code of ethics}
    \item[] Question: Does the research conducted in the paper conform, in every respect, with the NeurIPS Code of Ethics \url{https://neurips.cc/public/EthicsGuidelines}?
    \item[] Answer: \answerYes{}
    \item[] Justification: The research uses publicly available reanalysis (ERA5) and gauge-derived (PRISM) precipitation datasets, involves no human subjects, no personally identifiable information, and no models with foreseeable dual-use risks. All authors have reviewed the NeurIPS Code of Ethics and confirm conformance.

\item {\bf Broader impacts}
    \item[] Question: Does the paper discuss both potential positive societal impacts and negative societal impacts of the work performed?
    \item[] Answer: \answerYes{}
    \item[] Justification: Section 5.4 (Broader Impacts and Reproducibility) discusses positive applications --- flood-risk assessment, water-resource management, agricultural planning, and climate adaptation in hurricane- and atmospheric-river-affected regions --- and the principal negative-impact risks (over-reliance on the median output without consulting the calibrated quantile spread, and out-of-distribution use beyond the trained Florida/California/Texas Gulf-Coast domains and 1980--2013 ERA5--PRISM period), with a mitigation that emphasises calibrated uncertainty as the intended user interface.
    
\item {\bf Safeguards}
    \item[] Question: Does the paper describe safeguards that have been put in place for responsible release of data or models that have a high risk for misuse (e.g., pre-trained language models, image generators, or scraped datasets)?
    \item[] Answer: \answerNA{}
    \item[] Justification: The released artefacts are a regression model that maps ERA5 atmospheric reanalysis to gridded precipitation and a small set of \cvae{}-generated synthetic precipitation fields used purely for data augmentation. Neither poses misuse risks of the kind contemplated by this question (no language generation, no scraped or sensitive data, no image generation).

\item {\bf Licenses for existing assets}
    \item[] Question: Are the creators or original owners of assets (e.g., code, data, models), used in the paper, properly credited and are the license and terms of use explicitly mentioned and properly respected?
    \item[] Answer: \answerYes{}
    \item[] Justification: ERA5 reanalysis is provided by ECMWF/Copernicus Climate Change Service under the Copernicus license; PRISM precipitation data is provided by the PRISM Climate Group at Oregon State University and is freely available for research use. Both datasets are cited in the paper, and we use them within their stated terms (research/non-commercial). Baseline architectures (SRDRN backbone, residual blocks, and the BCSD/QM/SRGAN comparators referenced in Related Work) are credited to their original publications.

\item {\bf New assets}
    \item[] Question: Are new assets introduced in the paper well documented and is the documentation provided alongside the assets?
    \item[] Answer: \answerYes{}
    \item[] Justification: The new assets --- Q-SRDRN training code (TensorFlow/Keras), the IncrementBound layer and pinball loss implementation, the \cvae{} v5c PyTorch model and sampling pipeline, the calibration-diagnostics script, and the 83-sample FL \cvae{} augmentation set --- will be released on acceptance under an open-source license, with a README and configuration files documenting architecture, training command, hyperparameters, and dataset preparation steps (Section 5.4 Broader Impacts and Reproducibility).

\item {\bf Crowdsourcing and research with human subjects}
    \item[] Question: For crowdsourcing experiments and research with human subjects, does the paper include the full text of instructions given to participants and screenshots, if applicable, as well as details about compensation (if any)?
    \item[] Answer: \answerNA{}
    \item[] Justification: The paper does not involve crowdsourcing or research with human subjects. All data are publicly available reanalysis (ERA5) and gauge-derived gridded precipitation (PRISM).

\item {\bf Institutional review board (IRB) approvals or equivalent for research with human subjects}
    \item[] Question: Does the paper describe potential risks incurred by study participants, whether such risks were disclosed to the subjects, and whether Institutional Review Board (IRB) approvals (or an equivalent approval/review based on the requirements of your country or institution) were obtained?
    \item[] Answer: \answerNA{}
    \item[] Justification: The paper does not involve human subjects, so IRB approval is not applicable.

\item {\bf Declaration of LLM usage}
    \item[] Question: Does the paper describe the usage of LLMs if it is an important, original, or non-standard component of the core methods in this research? Note that if the LLM is used only for writing, editing, or formatting purposes and does \emph{not} impact the core methodology, scientific rigor, or originality of the research, declaration is not required.
    \item[] Answer: \answerNA{}
    \item[] Justification: LLMs were not used as a component of the core methodology. The Q-SRDRN architecture, IncrementBound layer, pinball loss, and \cvae{} augmentation pipeline are all conventional CNN/VAE-based models trained from scratch on ERA5 and PRISM. Any LLM usage was limited to writing assistance and code formatting, which the NeurIPS policy explicitly excludes from the declaration requirement.

\end{enumerate}